\documentclass{article}

\usepackage{microtype}
\usepackage{graphicx}
\usepackage{subfigure}
\usepackage{booktabs} 

\usepackage{tabularx}
\newcolumntype{Y}{>{\centering\arraybackslash}X}

\usepackage{caption}
\usepackage{nicefrac}       
\usepackage{listings}

\usepackage{tikz}
\usetikzlibrary{arrows}
\usetikzlibrary{bayesnet}
\usetikzlibrary{arrows}
\usetikzlibrary{trees}

\tikzset{
check/.style = {align=center, outer sep=0pt,
inner sep=0pt,label=east:\color{green}\scalebox{2}{$\checkmark$}},
cross/.style = {align=center, outer sep=0pt,
inner sep=0pt,label=east:\color{red}\scalebox{2}{$\times$}},
}

\usepackage{amsmath}
\usepackage{mathtools}
\usepackage{amssymb}
\usepackage{bm}
\usepackage{balance}
\newcommand{\R}{\mathbb{R}}

\newcommand{\X}{\mathcal{X}}

\newcommand{\Z}{\mathcal{H}}

\newcommand{\mat}[1]{\bm{#1}}

\DeclareMathOperator*{\argmin}{argmin}
\newcommand*{\diff}{\mathop{}\!\mathrm{d}}

\DeclareMathOperator{\p}{p}

\DeclareMathOperator{\Norm}{\mathcal{N}}

\DeclareMathOperator{\Uni}{\mathbb{U}}

\providecommand\given{}
\DeclarePairedDelimiterX{\Cond}[1]{(}{)}{
\renewcommand\given{%
  \nonscript\mkern2mu
  \delimsize\vert
  \nonscript\mkern2mu
  \mathopen{}
  \allowbreak}
#1
}

\makeatletter
\newcommand{\Fun}{\@ifstar\@sfun\@fun}
\newcommand{\@fun}[1]{#1\Cond}
\newcommand{\@sfun}[1]{#1\Cond*}
\makeatother

\newcommand{\Prob}{\p\Cond}

\newcommand{\Gaussian}{\Norm\Cond}

\newcommand{\Uniform}{\Uni\Cond}

\DeclarePairedDelimiterX{\KLdelim}[2]{(}{)}{%
  #1\mkern2mu\delimsize\|\mkern2mu#2%
}

\DeclarePairedDelimiterXPP{\Moment}[2]{#1}{[}{]}{}{
\renewcommand\given{%
  \nonscript\mkern2mu
  \delimsize\vert
  \nonscript\mkern2mu
  \mathopen{}
  \allowbreak}
#2
}

\DeclarePairedDelimiterX{\Set}[1]{\{}{\}}{

#1
}

\usepackage{csquotes}

\usepackage[disable]{todonotes}

\usepackage[accepted,nohyperref]{icml2020}

\icmltitlerunning{Modulating Surrogates for Bayesian Optimization}

\begin{document}

\newcommand{\us}{LGP}

\twocolumn[
\icmltitle{Modulating Surrogates for Bayesian Optimization}



\icmlsetsymbol{equal}{*}

\begin{icmlauthorlist}
\icmlauthor{Erik Bodin}{bristol}
\icmlauthor{Markus Kaiser}{siemens,munich}
\icmlauthor{Ieva Kazlauskaite}{bath}
\icmlauthor{Zhenwen Dai}{spotify}
\icmlauthor{Neill D. F. Campbell}{bath}
\icmlauthor{Carl Henrik Ek}{cam}
\end{icmlauthorlist}

\icmlaffiliation{cam}{University of Cambridge, United Kingdom}
\icmlaffiliation{bristol}{University of Bristol, United Kingdom}
\icmlaffiliation{bath}{University of Bath, United Kingdom}
\icmlaffiliation{siemens}{Siemens AG, Germany}
\icmlaffiliation{munich}{Technical University of Munich, Germany}
\icmlaffiliation{spotify}{Spotify Research, United Kingdom}

\icmlcorrespondingauthor{Erik Bodin}{erik.bodin@bristol.ac.uk}

\icmlkeywords{Machine Learning, Bayesian Optimization, Gaussian Processes, Robust, Challenging structures, ICML}

\vskip 0.3in
]



\printAffiliationsAndNotice{}  

\begin{abstract}
Bayesian optimization (BO) methods often rely on the assumption that the objective function is well-behaved,
but in practice, this is seldom true for real-world objectives even if noise-free observations can be collected.
Common approaches, which try to model the objective as precisely as possible, often fail to make progress by spending too many evaluations modeling irrelevant details.
We address this issue by proposing surrogate models that focus on the well-behaved structure in the objective function, which is informative for search, while ignoring detrimental structure that is challenging to model from few observations.
First, we demonstrate that surrogate models with appropriate noise distributions can absorb challenging structures in the objective function by treating them as irreducible uncertainty.
Secondly, we show that a latent Gaussian process is an excellent surrogate for this purpose, comparing with Gaussian processes with standard noise distributions.
We perform numerous experiments on a range of BO benchmarks and find that our approach improves reliability and performance when faced with challenging objective functions.
\end{abstract}

\section{Introduction}
\label{sec:introduction}
\begin{figure*}[h!]
    \centering
    \raisebox{-0.5\height}{\includegraphics[trim=15 15 0 0, clip, width=0.25\textwidth]{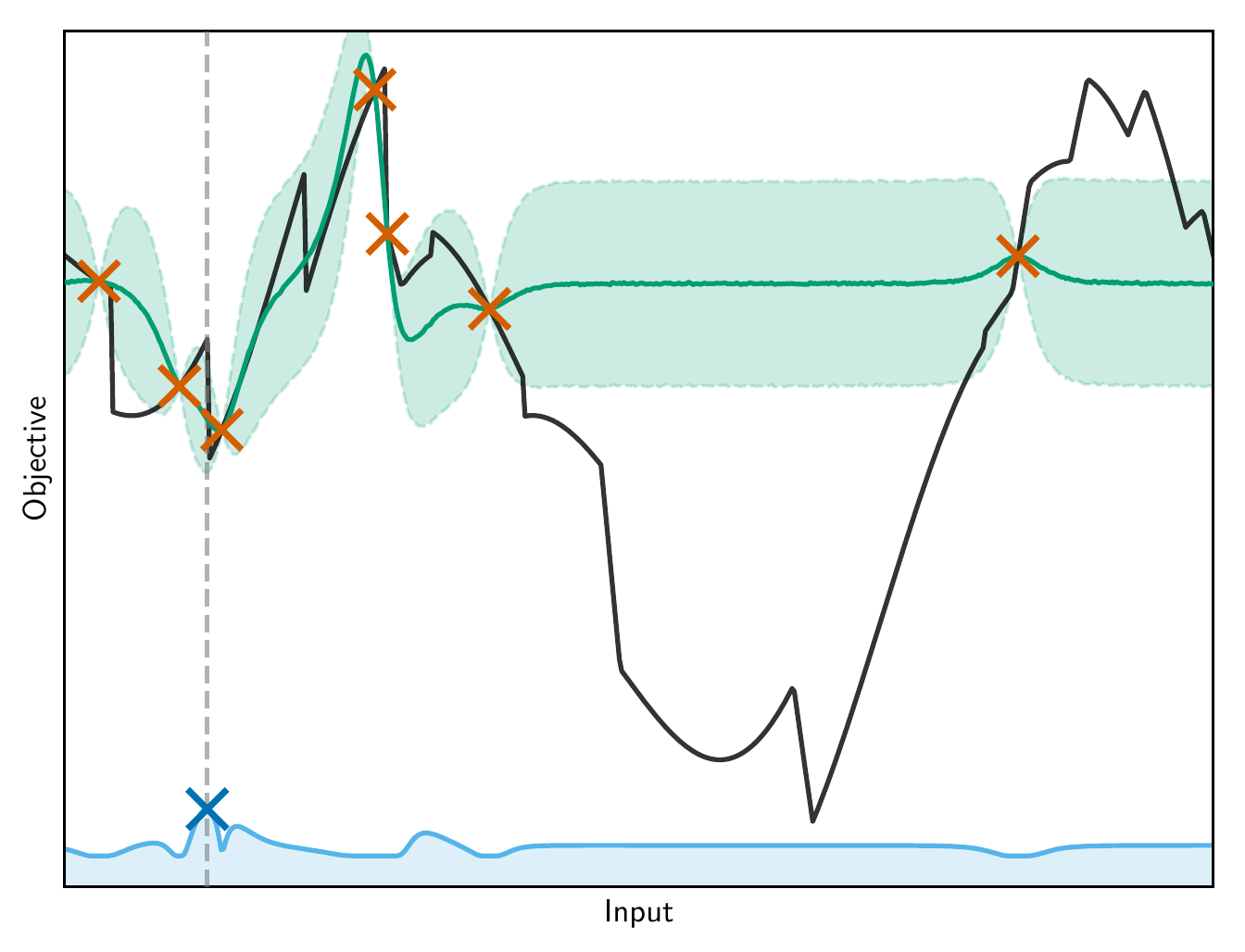}}
    \raisebox{-0.5\height}{\includegraphics[trim=15 15 0 0, clip, width=0.25\textwidth]{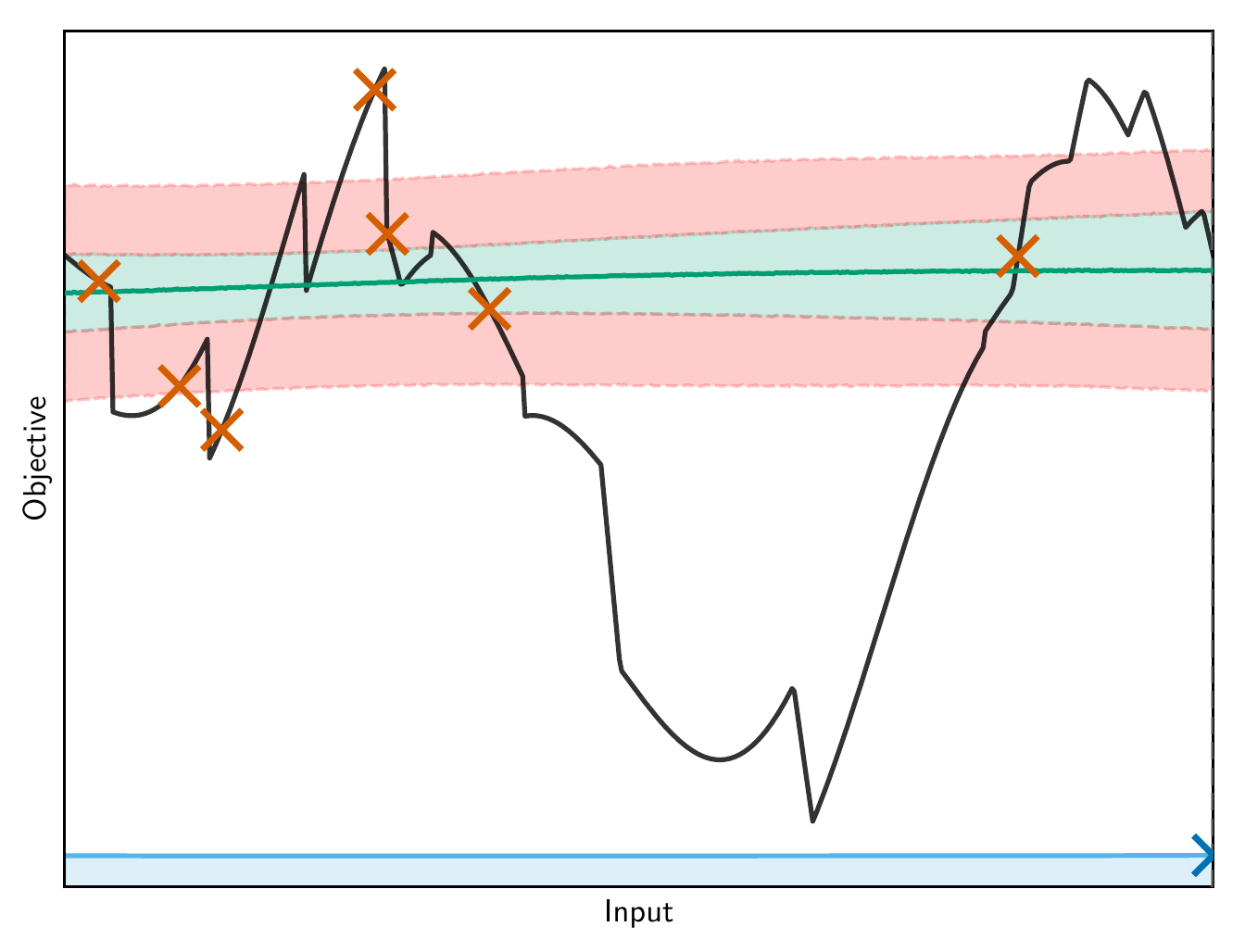}}
    \raisebox{-0.5\height}{\includegraphics[trim=15 15 0 0, clip, width=0.25\textwidth]{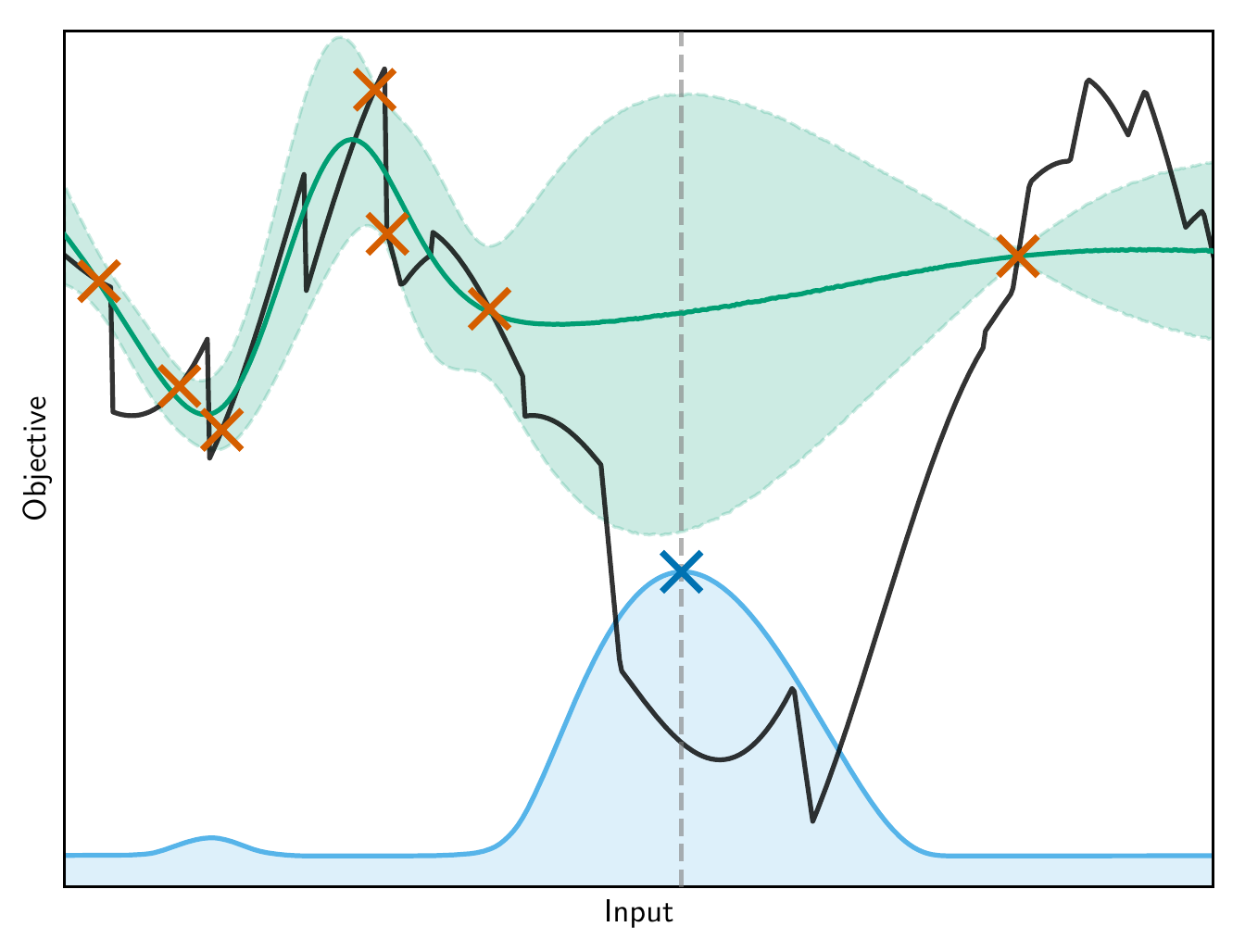}}
    \raisebox{-0.45\height}{{\includegraphics[width=0.20\textwidth]{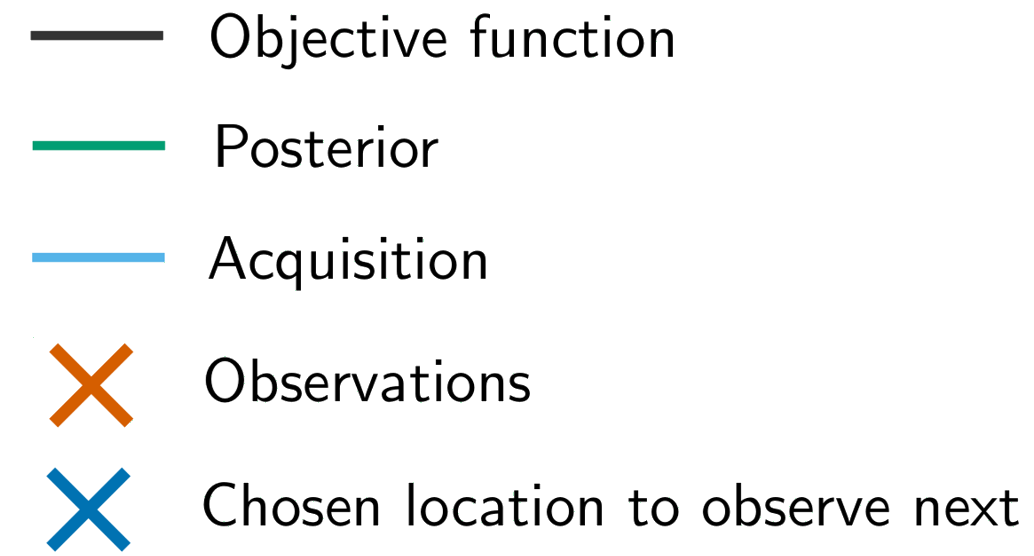}}}
    \caption{
    \label{fig:posterior}
    An illustrative example of the posterior surrogate function density obtained given observations of a 1D nonsmooth function
    using a noise-free GP, a GP with homoscedastic Gaussian noise, and using a LGP model.
    The posterior belief for the noise-free and homoscedastic GP surrogates results in the EI acquisition function,
    shown in blue,
    making poorly informed decisions for the next query.
    In contrast, the LGP using our proposed setup is able to reduce the influence of the rapid oscillations that do not match the GP prior by explaining part of the variation using the latent input.
    As a result, the acquisition function can utilize a confidently discovered global trend to increase the efficiency of the search.
    In this example, $\sigma_h$ is set to $\frac{1}{50}$ of the domain range to allow ignorance of oscillations at that scale.
    }
\end{figure*}
Bayesian optimization (BO)~\cite{snoek2012practical} is a method for finding the optimum of functions that are unknown and expensive to evaluate.
By fitting a surrogate model to the samples of an unknown objective, the BO procedure iteratively picks the new samples of the objective believed to be the most informative about where the optimum is located.

Model misspecification has significant negative implications for any machine learning tasks.
This is especially true for sequential decision making tasks such as BO, where the model is used not only to locate the optimum based on the collected data but also to decide where to collect data for future decisions.
If the surrogate model is misspecified, it is likely to acquire samples that are less informative about the optimum, which will lead to a less efficient optimization.
Therefore the quality of the surrogate model is essential to achieve both efficient and reliable results.

Many works have been done towards avoiding model misspecification in the surrogate model for BO, such as handling non-stationary objective functions with warpings~\cite{snoek2014input}, tree-structured dependencies in the search space~\cite{jenatton2017bayesian}, and searching the optimum from piecewise comparisons~\cite{gonzalez2017preferential}.
Comparing with the Gaussian process (GP) regression model in the standard BO setting, these methods avoid model misspecification in real-world problems by using more sophisticated surrogate models that are suitable for the corresponding problems.
Bayesian inference with more sophisticated surrogate models will often require additional data to reduce uncertainty and confirm beliefs, because it considers more possibilities.
Importantly the ultimate goal of BO is to find the optimum, not to model the unknown objective as precisely as possible.
In practice, this means that using a surrogate with high complexity might perform worse compared to a simpler class
even if the former contains the true objective function.

Instead of building a complex surrogate model with minimal model misspecification,
we propose an alternative approach which allows trading off accuracy in modeling the objective with efficiency of capturing informative structures from small amounts of data.
For example, we observe that structures such as local oscillations and discontinuities are less important to capture for the purposes of BO.
Such details often require a lot of data to be closely captured in a surrogate model but do not help the search for the optimum,
unless the search reaches the last stage of pinpointing the exact location of the optimum.
To ignore these details, we associate an independent random \emph{input} variable with every evaluation of the unknown function.
As the random variables associated with new evaluations are conditionally independent of the posterior random variables associated with observed data given the function,
this is referred to as \emph{irreducible uncertainty}.
Such variables are similar to the noise variables in regression models, which are used to capture measurement noise and the data variance that cannot be attributed to the input variables.
In contrast to noise variables for noisy outcomes,
where there is irreducible uncertainty about the data, there is now irreducible uncertainty in the model of the function.

We propose to use the surrogate models that are specified over well-behaved approximations of the objectives,
which can be more useful for the search of the optimum (see Figure~\ref{fig:posterior}), augmented with flexible ``noise" distributions to treat the nuisance parameters.
We will demonstrate that, using the same function approximation, a surrogate model with a more flexible nuisance parameter distribution is more robust against challenging structures.
In this paper we focus on noise-free objectives with complicated, oscillatory or discontinuous structures.
In particular, we propose to use a Latent Gaussian process (LGP)~\cite{pfingsten2006nonstationary,wang12_gauss_proces_regres_with_heter,yousefi_unsupervised_2016,bodin_latent_2017} as the surrogate model due to its flexible nuisance parameter distribution and show that it outperforms the surrogate models with less flexible distributions such as GPs with additive likelihoods.
LGP allows us to disentangle the complicated structures a GP surrogate struggles to model while highlighting important structures.

Our main contributions are:
\begin{itemize}
    \item We propose to address challenging objective functions for BO by using a distribution in the surrogate model to explain structure that is challenging to model with few observations.
    \item We propose to use latent Gaussian processes (LGP) as surrogate models, which support non-stationary and non-Gaussian residuals.
    \item With experiments on multiple BO benchmarks, we show that our method significantly outperforms existing approaches.
\end{itemize}

\section{Modulating Surrogates}
\label{sec:modulated_objectives}

Let $f : \X \to \R$ be an unknown, noise-free objective function defined on a bounded subset $\X \subset \R^Q$.
The goal of BO is to solve the global optimization problem of finding
\begin{equation}
    \mat{x}_{\text{min}} = \argmin_{\mat{x} \in \X} f(\mat{x}).
\end{equation}
In real world problems,
the objective function is often not a well-behaved function and a suitable model is difficult to specify.
Instead of applying an automated model selection method~\cite{malkomes2018automating},
we propose to model only the essential structure of the objective function that is well-behaved and leave the rest of the function details to be absorbed in a noise distribution.

We consider the family of objective functions $f$ that can be represented as a composition of a well-behaved function and another arbitrary, latent function capturing the challenging details, i.e.
\begin{equation}
    \label{eq:well_behaved}
    f(\mat{x}) := g(\mat{x}, \mat{h}), \quad \mat{h}:=h(\mat{x}),
\end{equation}
where $g$ is a well-behaved function that can be nicely modeled by a surrogate model of choice,
which is a Gaussian process (GP) in this paper,
and the vector-valued function $h(\mat{x})$ encodes the structures which the surrogate model struggles to capture.
In general this composition allows for complicated interactions between $\mat{x}$ and $\mat{h}$, producing complicated realizations of the function
which is observed through data.
A simple, special case of a function composition is additive structure
\footnote{Note that in the additive case, $h(\mat{x})$ must match the output in shape, i.e.~be one-dimensional.},
i.e.\ $f(\mat{x}) = g(\mat{x}) + h(\mat{x})$.

Instead of modeling $h(\mat{x})$ as part of the surrogate model,
we propose to \textit{ignore} the structure of the objective function in $h(\mat{x})$ by replacing $h(\mat{x})$ with a random variable $\mat{h}$ per data point.
The random variables $\mat{h}$ for different data points  are independent among each other.
The objective function becomes a function of two variables $g(\mat{x}, \mat{h})$,
in which $\mat{h}$ is a random variable which explain the data variance that cannot be explained by $\mat{x}$.
In this paper, we use a normal distribution for the prior of $\mat{h}$, $\mat{h} \sim \mathcal{N}(0,  \mathbb{I})$.
Note that, although the distribution of $h(\mat{x})$ induced by the data distribution for $\mat{x}$ may not be zero-mean and unit-variance, it is easy to reformulate it as a linear transformation of a normal distribution with zero-mean and unit-variance and the resulting linear transformation can be absorbed into the function $g$.
For further details on the definition, see the supplement.

With the above formulation, a BO method can be developed by constructing a model of the well-behaved function $g$
and a model of $h$.
At each step of the BO optimization, a set of input and output pairs of the objective function has been collected, denoted as $\mat{X} = (\mat{x}_1, \ldots, \mat{x}_N)^\top$ and $\mat{F} = (\mat{f}_1, \ldots, \mat{f}_N)^\top$.
The output $\mat{F}$ denotes the noise-free observations of the objective function.
The Bayesian inference of the model aims at inferring the posterior distribution
\begin{equation}
    \Prob{ \mat{H}, \mat{\theta} \given \mat{X}, \mat{F}}  \propto \Prob{\mat{F} \given \mat{X}, \mat{H}, \mat{\theta} }\Prob{\mat{H}}  \Prob{\mat{\theta}}
\end{equation}
where $\mat{\theta}$ are the hyperparameters of the surrogate model and $\mat{H} = (\mat{h}_1, \ldots, \mat{h}_N)^\top$ is the concatenation of the nuisance parameters associated with the individual data points.
The location of the next evaluation is determined according to an acquisition function,
which uses the predictive distribution $\Prob{\mat{f}_* \given \mat{x}_*,  \mat{X}, \mat{F}}$ of the surrogate model,
\begin{align}
    \label{eq:pred_dist}
    \begin{split}
        \MoveEqLeft\Prob{\mat{f}_* \given \mat{x}_*,  \mat{X}, \mat{F}} =
        \int \Prob{\mat{f}_* \given \mat{x}_*, \mat{h}_*, \mat{X}, \mat{F},\mat{H}, \mat{\theta}}\\
        &\qquad\Prob{\mat{H}, \mat{\theta} \given \mat{X}, \mat{F}} \Prob{\mat{h}_*} \diff{\mat{H}}\diff{\mat{\theta}}\diff{\mat{h}_*},
    \end{split}
\end{align}
where $\mat{x}_*$ is the input of the prediction and $\mat{f}_*$ is the noise-free observation at the location $\mat{x}_*$.
The predictive distribution of the latent variable $\Prob{\mat{h}_*}$ associated with new evaluations is as of the i.i.d.~assumption
equal to the prior.
As such $\Prob{\mat{h}_*}$ contains model uncertainty \emph{irreducible} by the active sampling loop,
which we suggest to ignore via augmentation,
see the supplement for details.

With the predictive distribution Eq.~\ref{eq:pred_dist},
the expectation of the acquisition function is derived as
\begin{align}
    \begin{split}
        \label{eq:utility}
        \MoveEqLeft[1]\alpha(\mat{x}_\ast) =
        \int
        \mathbb{U}(\mat{f}_\ast, \mat{x}_\ast, \mat{X}, \mat{F}) p(\mat{f}_\ast | \mat{x}_\ast, \mat{X}, \mat{F}) \diff{\mat{f}_\ast},
    \end{split}
\end{align}
where the acquisition function of choice is denoted $\mathbb{U}$.
Note that the predictive distribution due to the marginalization over $\mat{H}$ and $\mat{\theta}$ generally has a complicated form
and that the above integral often requires approximate methods.

\section{Latent GP surrogates and other choices}
\label{sec:latent_gp_surrogate}

\begin{figure*}[t]
    \centering
    \includegraphics[trim={0.6cm 0.6cm 0 0},clip, width=0.19\textwidth]{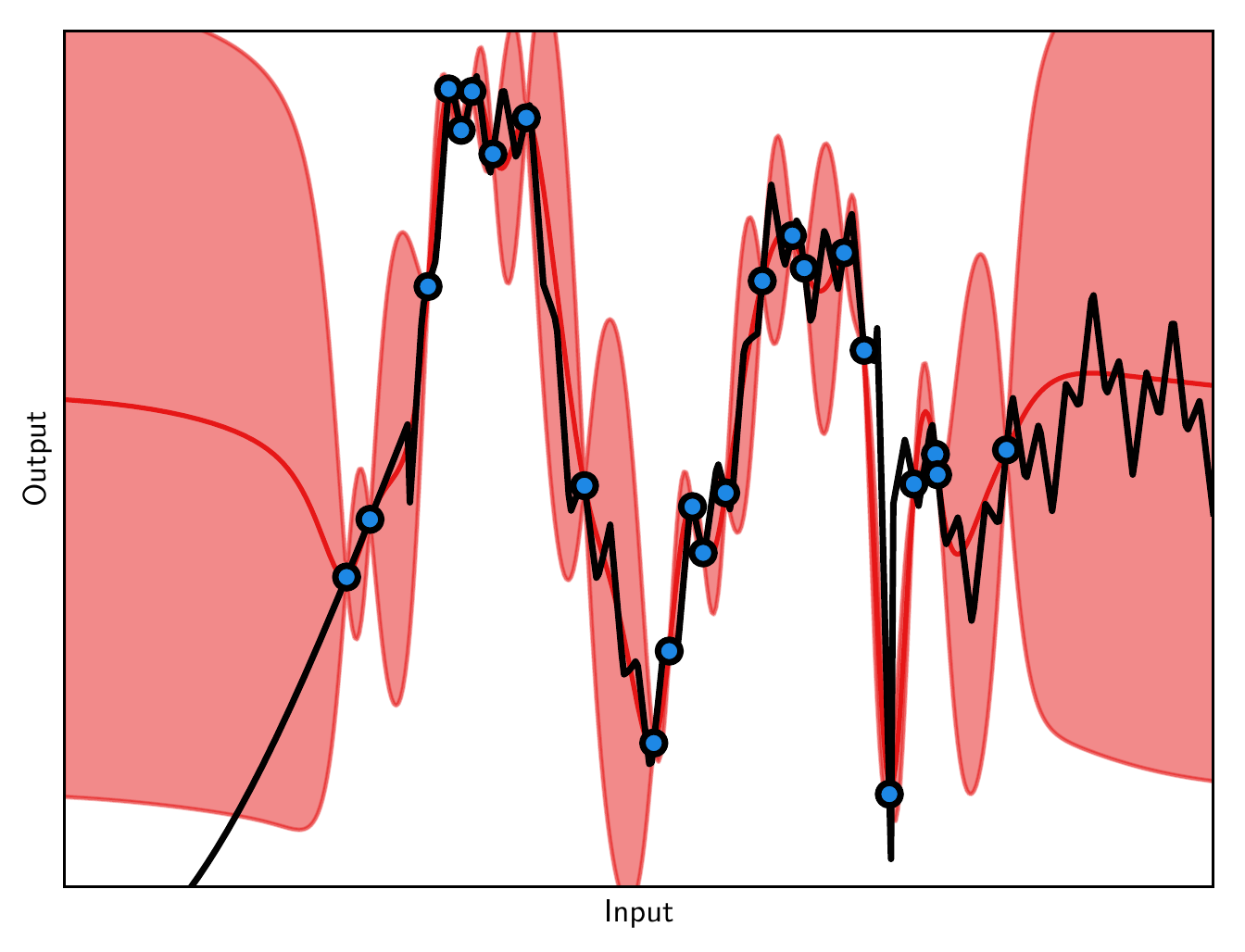}
    \includegraphics[trim={0.6cm 0.6cm 0 0},clip, width=0.19\textwidth]{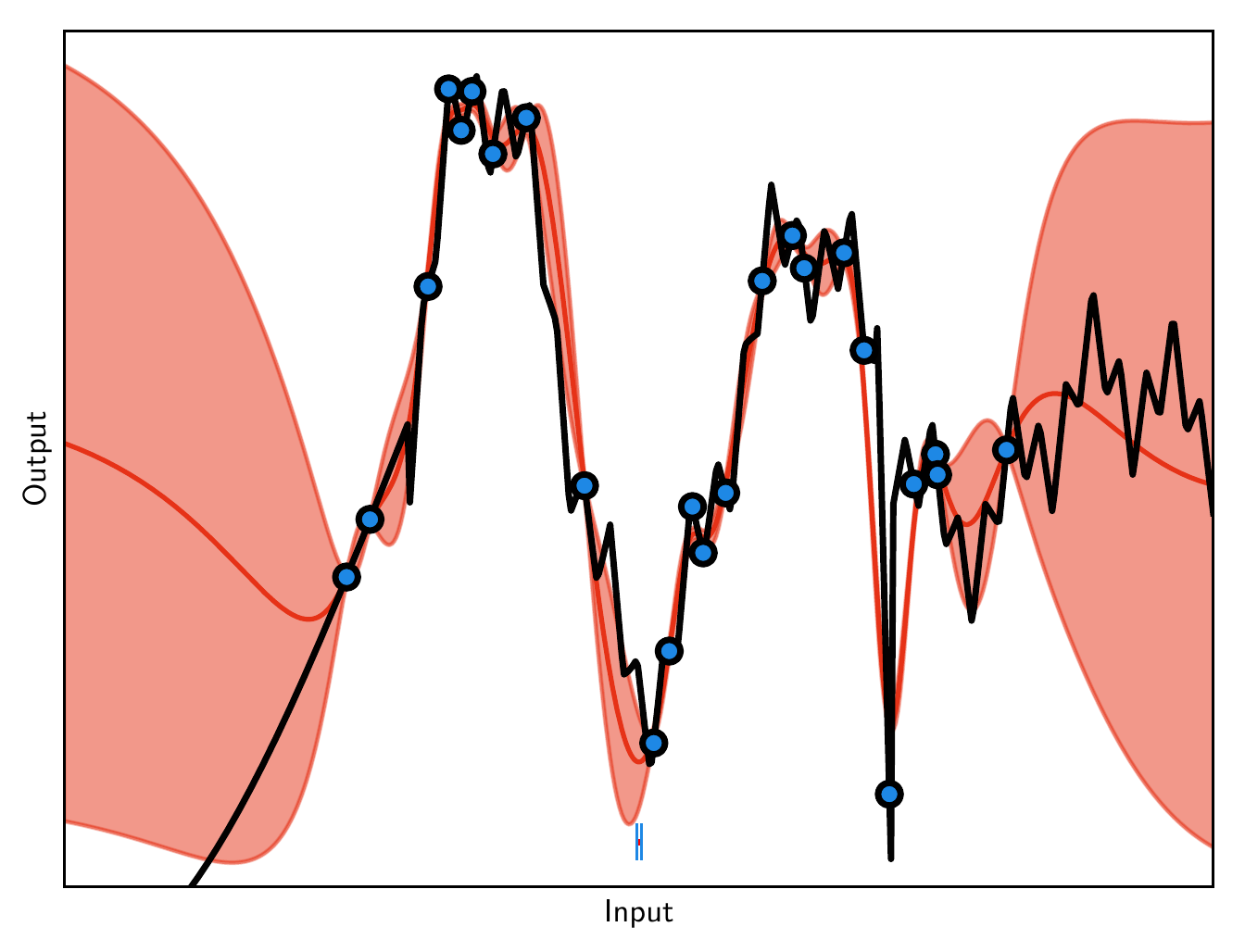}
    \includegraphics[trim={0.6cm 0.6cm 0 0},clip, width=0.19\textwidth]{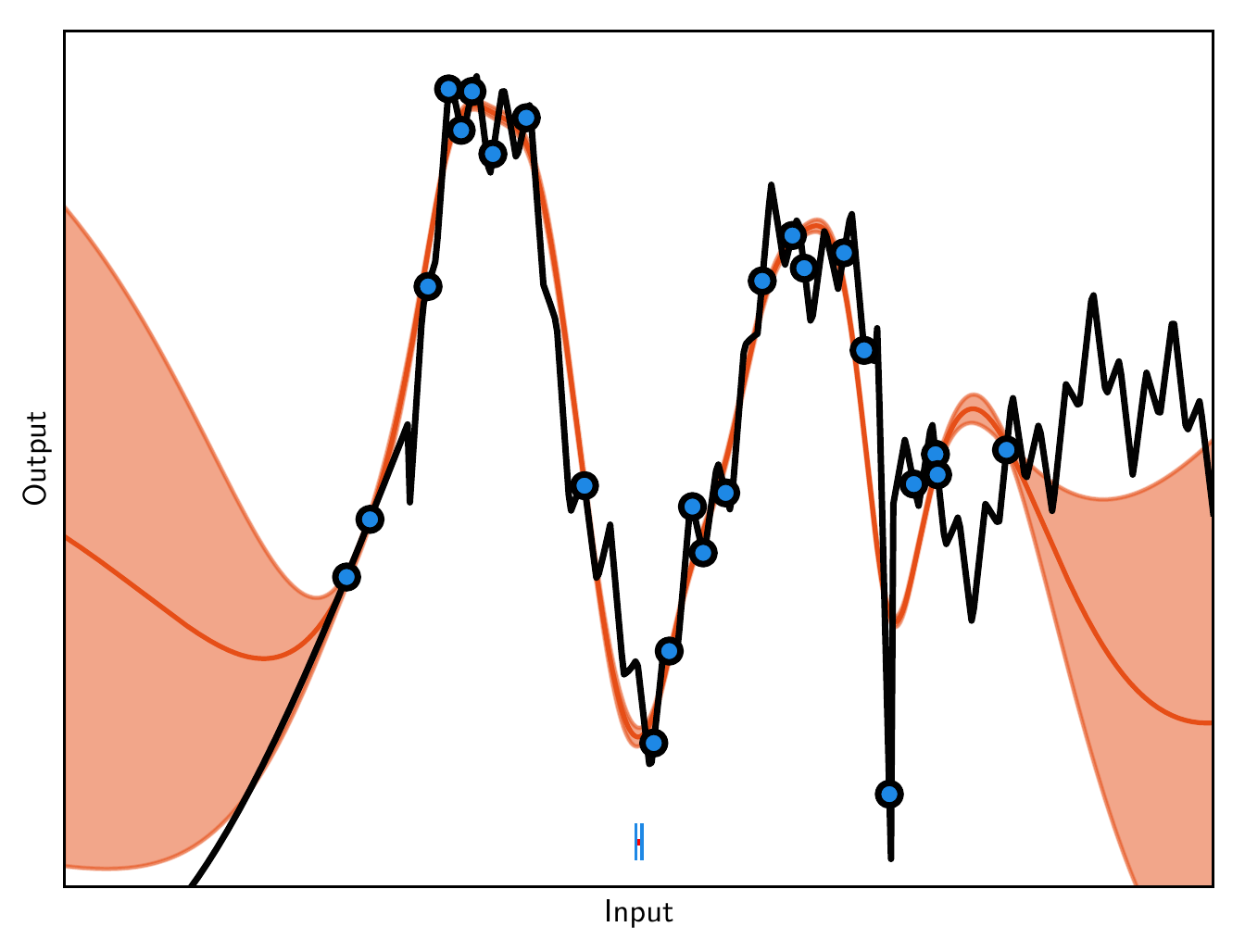}
    \includegraphics[trim={0.6cm 0.6cm 0 0},clip, width=0.19\textwidth]{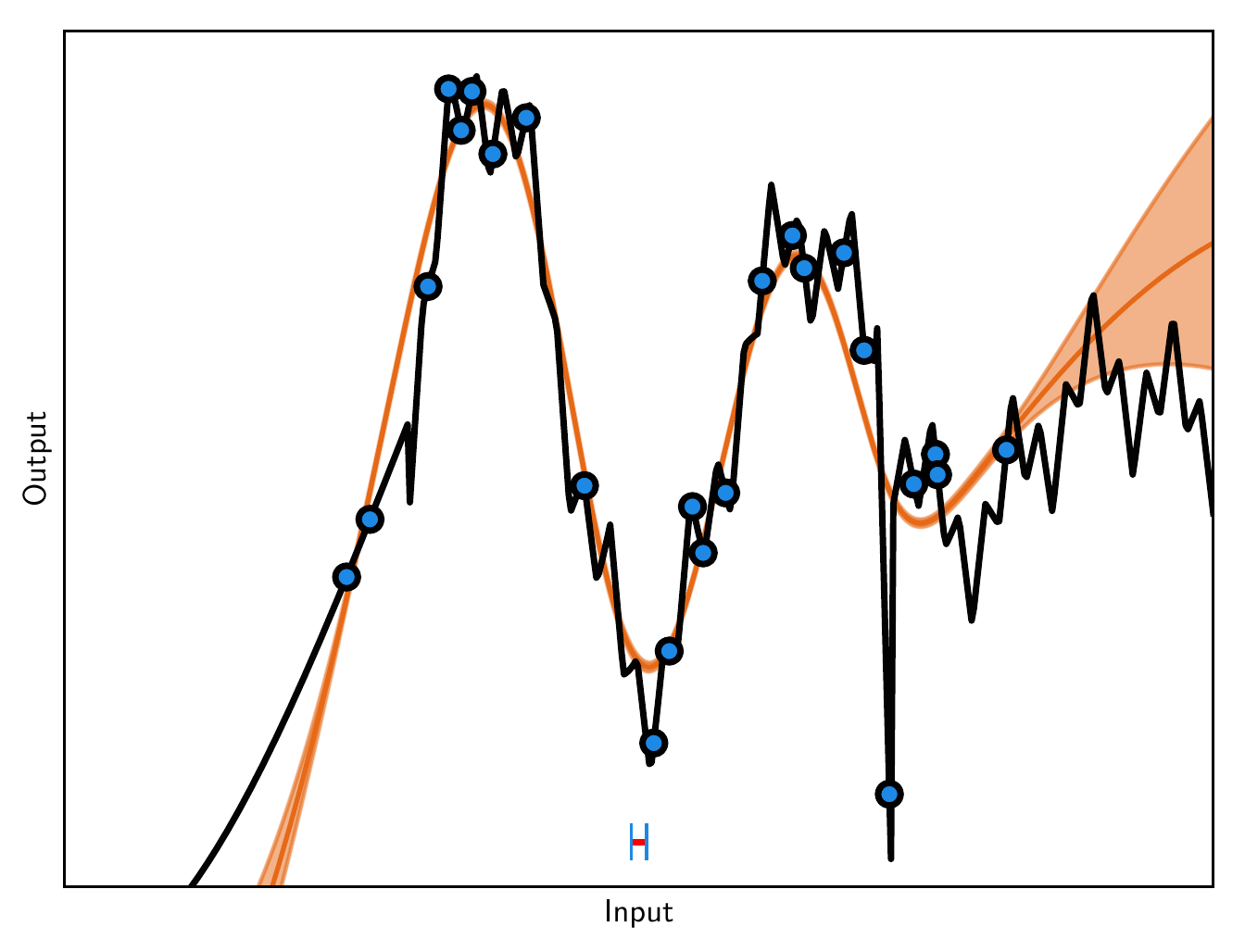}
    \includegraphics[trim={0.6cm 0.6cm 0 0},clip, width=0.19\textwidth]{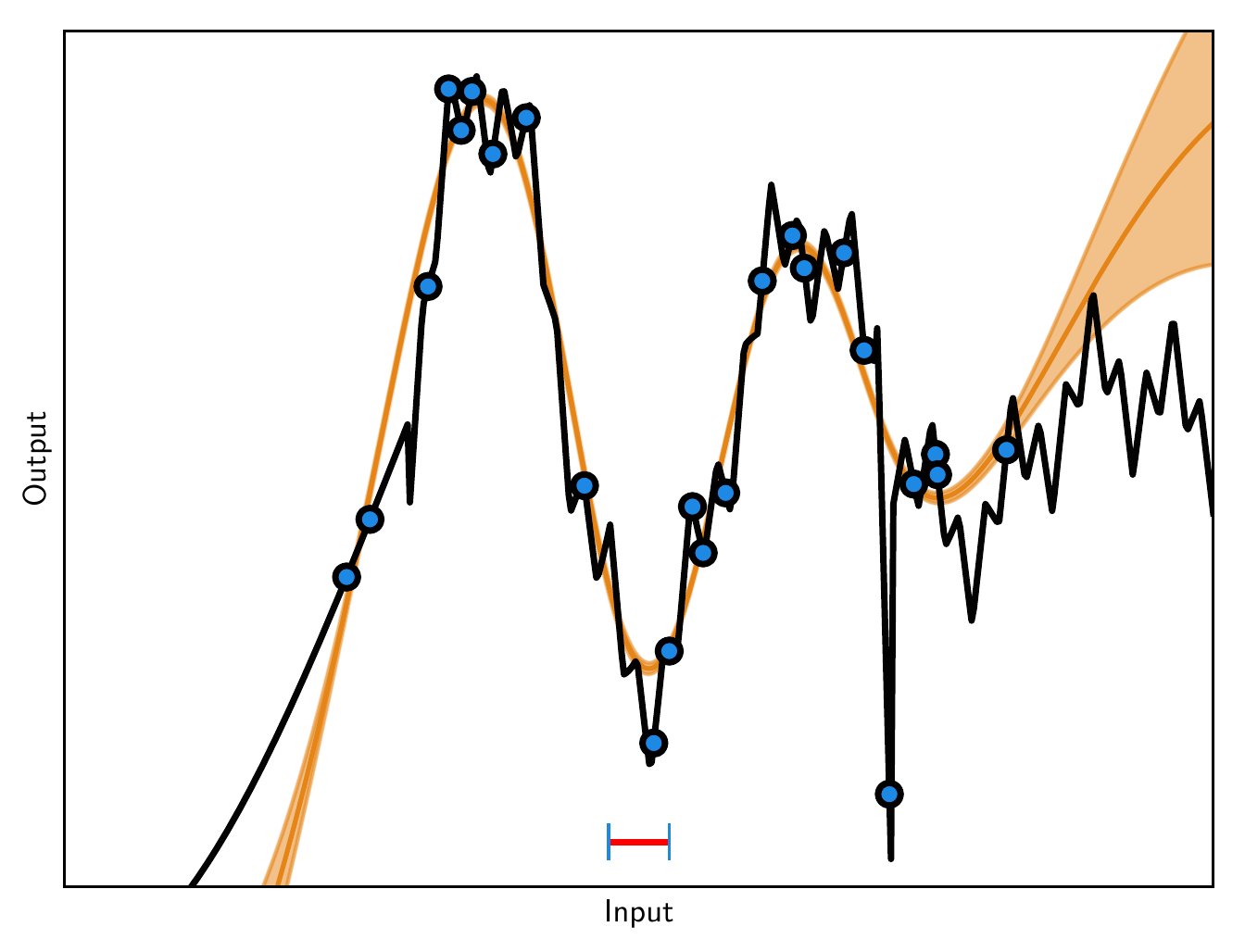}\\
    \hspace{-0.01cm}
    \includegraphics[trim={0.6cm 0.6cm 0 0},clip, width=0.19\textwidth]{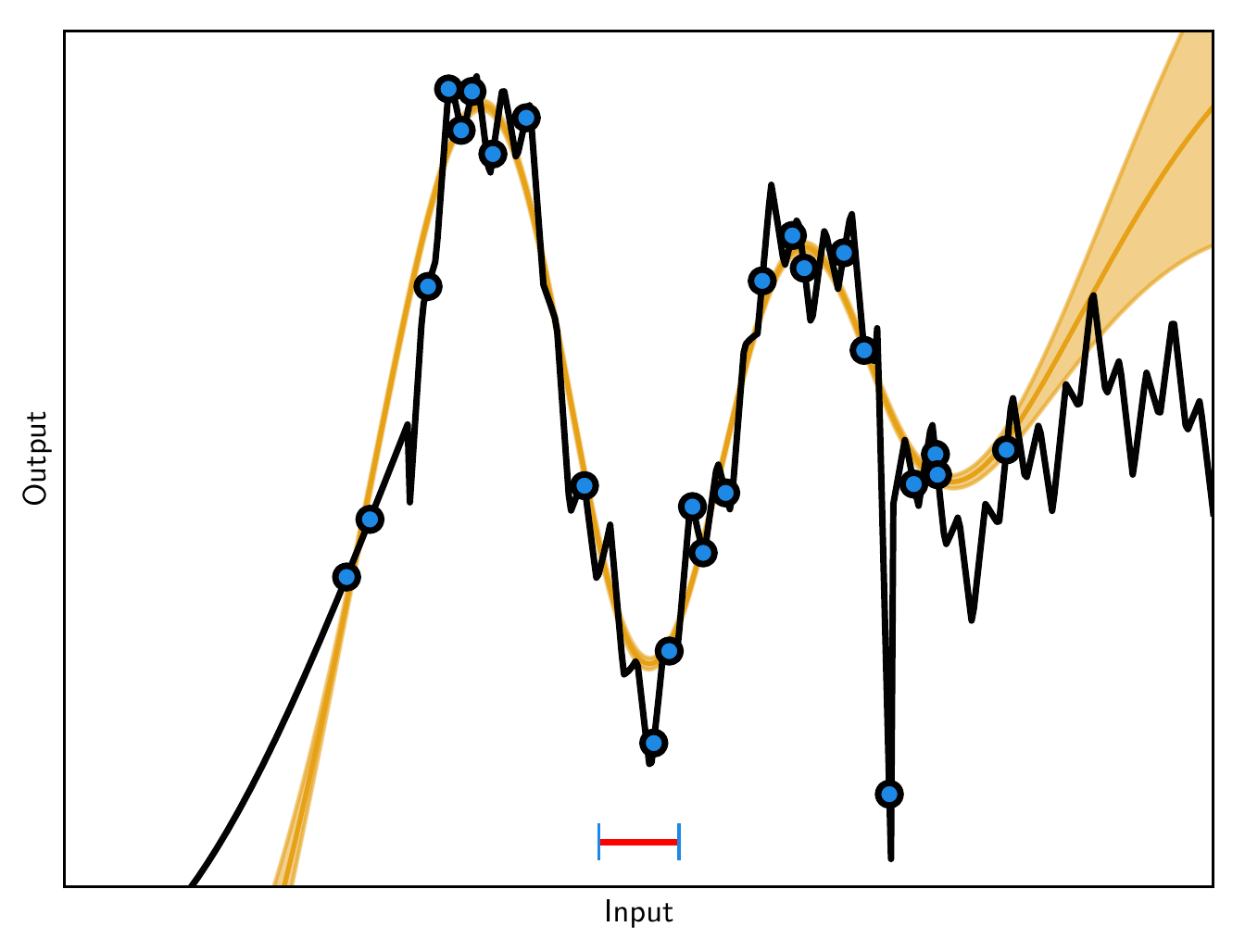}
    \includegraphics[trim={0.6cm 0.6cm 0 0},clip, width=0.19\textwidth]{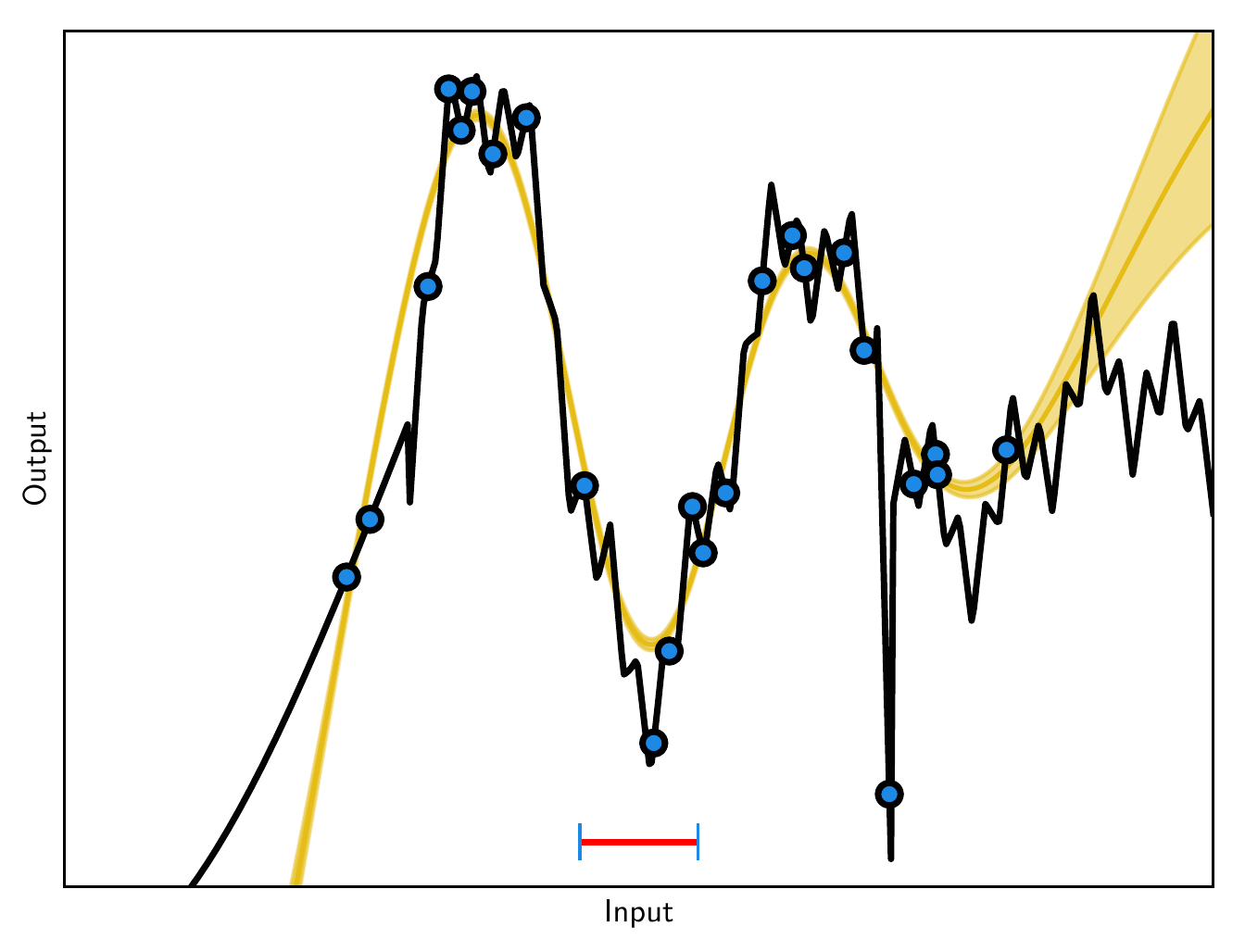}
    \includegraphics[trim={0.6cm 0.6cm 0 0},clip, width=0.19\textwidth]{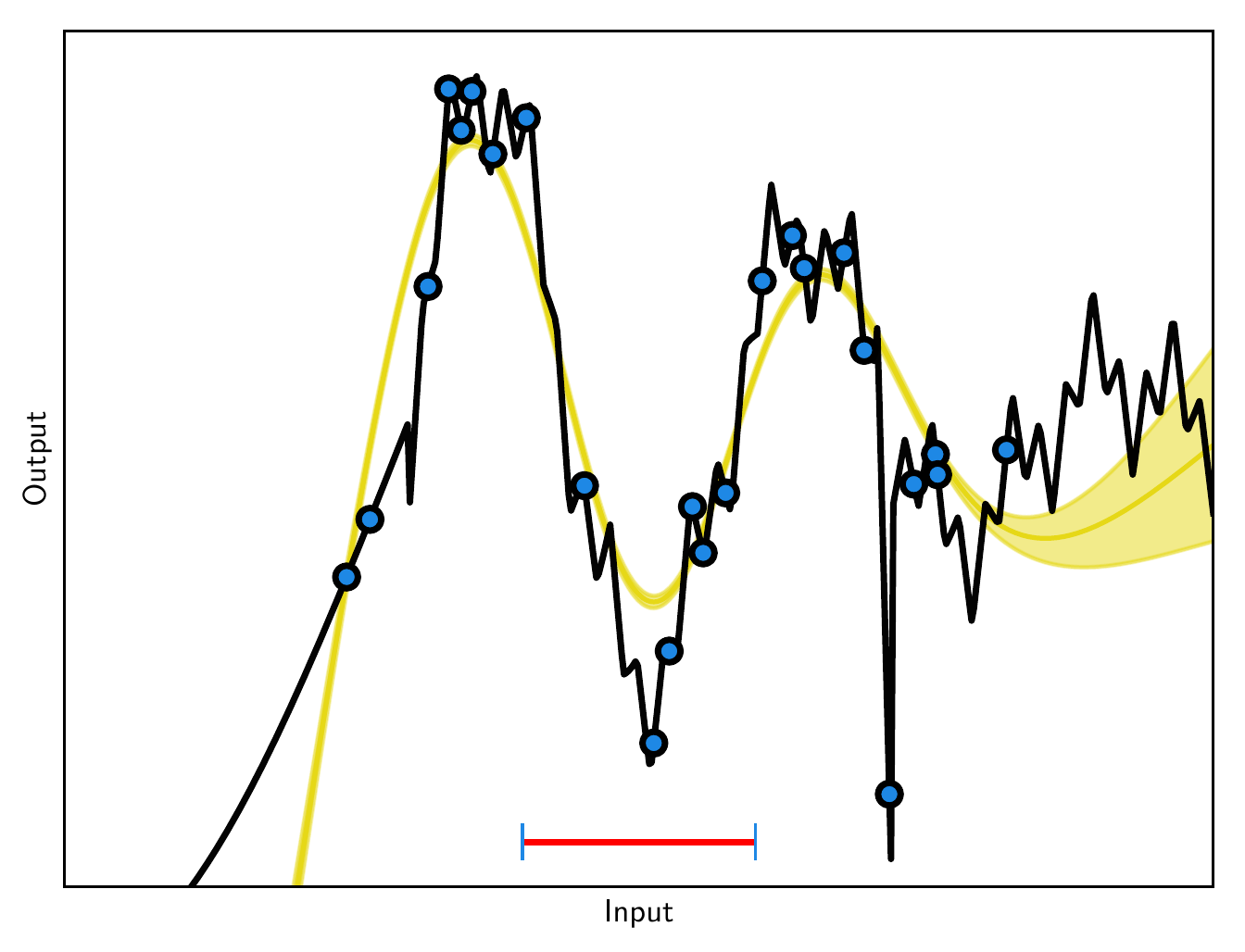}
    \includegraphics[trim={0.6cm 0.6cm 0 0},clip, width=0.19\textwidth]{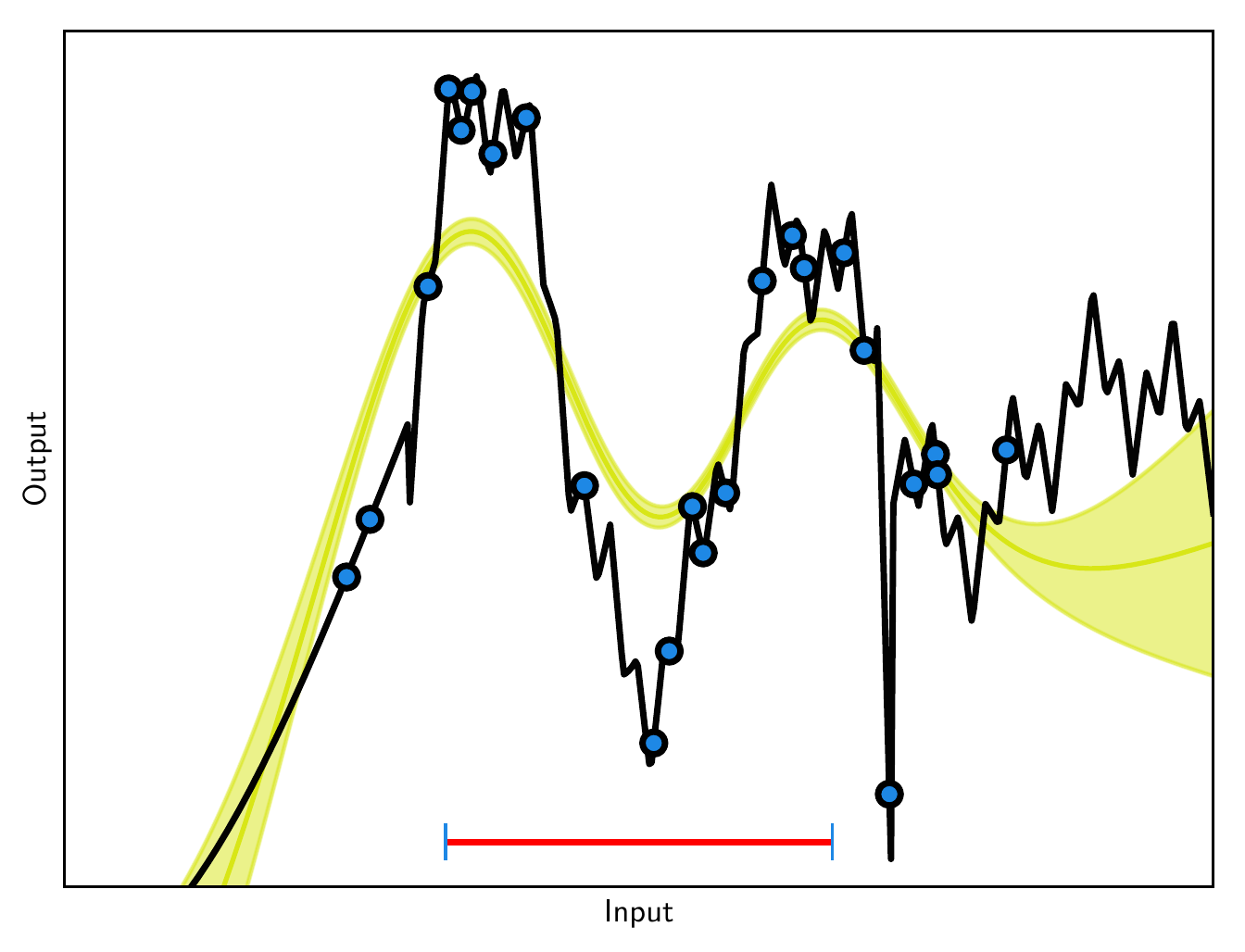}
    \includegraphics[trim={0.6cm 0.6cm 0 0},clip, width=0.19\textwidth]{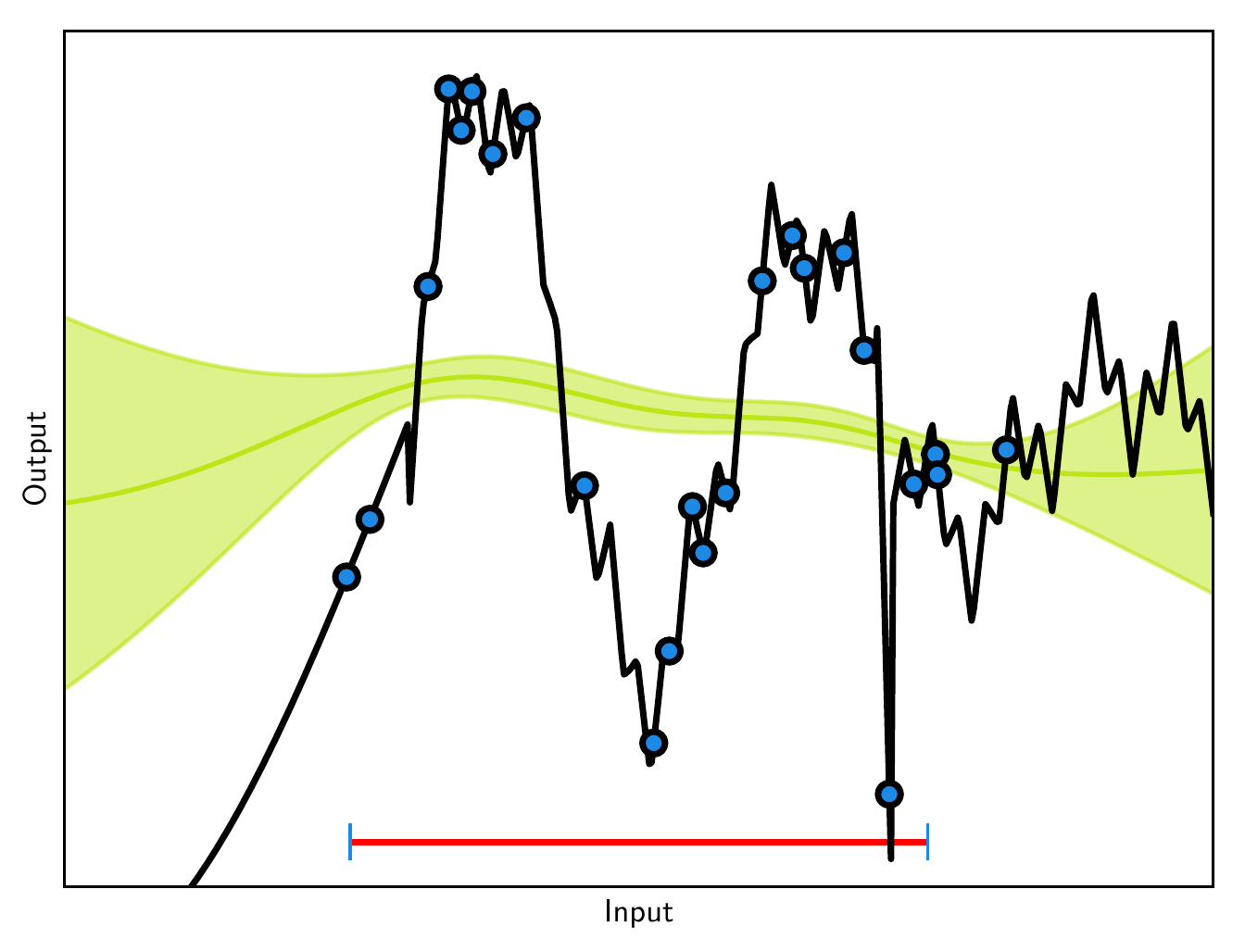}
    \caption{
    \label{fig:varying_z_sigma}
    Input-related invariance.
    Each plot is showing the resulting modulated function posterior using the LGP model and setting $\sigma_h$ in $p(\mat{h})$ to a size corresponding to the
    red line at the bottom of respective plot.
    The posterior is shown with mean and two standard deviations.
    The true function is shown in black.
    Note how the value of the prior $\sigma_h$ sets the scale in relation to $\X$ on how much detail is ignored.
    A connection can be made to low-pass filtering of higher frequencies, but where the filter varies between observations as of the posterior and where each filter is implicitly determined
    by fit to the function prior.
    }
\end{figure*}

In the previous section we presented the BO formulation.
We will now proceed to implement the formulation, and address the choice of surrogate model for the function (Eq.~\ref{eq:well_behaved}).

\textbf{Additive noise model.}
As briefly mentioned in the previous section, a simple case of the composition (Eq.~\ref{eq:well_behaved}) is an additive structure, $f(\mat{x}) = g(\mat{x})+h(\mat{x})$.
Following the process of replacing $h(\mat{x})$ with the random variable $\mat{h}$, the resulting surrogate model of the objective function is
\begin{equation}
    f(\mat{x}) = g(\mat{x}) + h, \quad h \sim \mathcal{N}(0, \sigma(\mat{x})^2),
\end{equation}
where the variance of $h$ is assumed to be $\sigma^2$ in order to adapt to the value range of $f$.
With a GP surrogate model for $g$, the above model recovers the GP regression model with the Gaussian likelihood.

A typical choice in the above model is to assume $\sigma^2$ to be constant, leading to a homoscedastic model.
A limitation of noise variances being the same across all the datapoints is that it limits the capability of the model in terms of absorbing irregular variance.
A straight-forward extension of the above model is the GP with heteroscedastic noise, in which the noise variance $\sigma^2$ is allowed to be different among data points \cite{goldberg1998regression,lazaro2011variational}.
Another choice could be specifying a GP prior for $h(\mat{x})$ and thus recover an additive GP model for $f$~\cite{bernardo1998regression,duvenaud2011additive}.
Other available choices for an additive noise model include Student's t-distribution~\cite{jylanki2011robust}, Laplace~\cite{kuss2006gaussian} or
mixture of Gaussian likelihoods as~\cite{kuss2006gaussian, stegle2008gaussian, naish2008robust} where~\cite{naish2008robust} considers the heteroscedastic case.

\textbf{Latent Gaussian process.}
A major limitation of the additive noise models in general is the inability to capture the interaction between the input $\mat{x}$ and the noise $\mat{h}$.
Another choice that produces a more flexible noise distribution is to introduce additive noise in the \emph{input} of a GP~\cite{mchutchon2011gaussian,girard2003gaussian,girard2004approximate}.
This would correspond to the case of $f(\mat{x}) = g(\mat{x} + \mat{h})$.
A further more general case of the proposed methodology is to allow non-linear interactions between the random variable $\mat{h}$ and $\mat{x}$.
This can be formulated as
\begin{equation}
    f(\mat{x}) = g(\mat{x}, \mat{h}),\quad g \sim \mathcal{GP},\quad \mat{h} \sim \mathcal{N}(0, \mathbb{I}). \label{eqn:lgp}
\end{equation}
This formulation aligns with the general assumptions proposed in the previous section.
In particular, the well-behaved function $g$ is assumed to follow a GP prior distribution,
and the random variable derived from the challenging details of the objective function $\mat{h}$ feeds directly into the GP surrogate model.
This allows for an arbitrary interaction between $\mat{h}$ and $\mat{x}$, as specified by the covariance function.
The introduction of the random variable $\mat{h}$ in the input results in a flexible noise distribution, as the GP model can warp the normal distribution of $\mat{h}$ into a sophisticated distribution and allow non-linear interactions between $\mat{h}$ and $\mat{x}$.
This GP model in (\ref{eqn:lgp}) is also known as a latent Gaussian process (LGP)~\cite{pfingsten2006nonstationary,wang12_gauss_proces_regres_with_heter,yousefi_unsupervised_2016,bodin_latent_2017}, which is developed for regression with heteroscedastic noise and non-Gaussian residuals.
The non-Gaussian marginals arise as a consequence of the latent covariates and their nonlinear transformation through the covariance function.

\paragraph{Function modulation via $\Z$}
If we assume a stationary kernel over the product space $\X \times \Z$, a constant $\mat{h}_n$ for all observations can be interpreted as the $\Z$ subspace having no influence.
This is due to the stationary property of the kernel, where covariances are determined only by the distances between points.

With everything else held constant, if an observation is moved away from other observations in the $\Z$ space, the covariances between that observation and the others are reduced.
Similarly, if the length scale in $\X$-direction is shortened, the covariances between that observation and the others can be equally reduced, but that also reduces the covariances between \emph{all other} observations due to the global influence of the hyperparameter.

Structures in the data could be explained solely by reducing the $\X$-direction length scale adequately.
In that case, evaluating the posterior at $\mat{h}_\ast = \mat{0}$ would yield exactly the posterior of a standard GP.
Conversely, structures could be explained solely as observations being adequately far from each other in the $\Z$ space while maintaining a longer $\X$ length scale.
Evaluating the posterior at $\mat{h}_\ast = \mat{0}$ then yields a posterior that is both influenced by the longer length scale and which has lower covariances with the data, effectively producing a posterior over smoother functions.
If the posterior inputs $\mat{H}$ are sufficiently far away with respect to the $\Z$-direction length scale,
all data variation will be captured in $\Z$ and the posterior of the function at $\mat{h}_\ast = \mat{0}$ will in effect ignore the data.

The posterior weighting over this range of solutions is determined by the trade-off between the GP function prior and the prior of the latent inputs.
As such, by controlling this trade-off, we can control properties of structures to be ignored and the ones to be used for search (see Figure~\ref{fig:varying_z_sigma}).
Important to note is that the mentioned data ignorance effect affects \emph{individual} data points via the posterior of the corresponding latent input $\mat{h}_n$,
which is influenced by the local and global fit of the function prior.

\paragraph{Reparameterization of LGP for ease of specifying the modulation prior}
In a BO setting, some prior knowledge about what constitutes a significant change in the input space is often available.
We would like to specify a joint prior of the GP and the latent inputs to ignore structures at the appropriate scale.
In order to do this, we (re)-parametererise it in the following way.
We set the lengthscale in the $\Z$-direction to be the same as in the $\X$-direction and parameterize the latent input prior as $\mathcal{N}(\mat{h}_n|\mat{0}, \sigma_h^2\mathbb{I})$ instead of a unit Gaussian.
There is an equivalence between parameterizing $\sigma_h$ or setting this trade-off via a separate lengthscale for $\Z$ as in~\cite{wang12_gauss_proces_regres_with_heter}.
However, by using the above parameterization there is a direct correspondence in covariance reduction from moving an observation in $\Z$ as in $\X$, and the prior for the latent inputs can be interpreted as a prior over the coarseness of the function we wish to exploit for search.
As such, intuitions about the scale in $\X$ directly translate into the parameterization of the prior.
See Figure~\ref{fig:varying_z_sigma} for a visualization of how changing $\sigma_h$ in the prior affects the modulated function posterior.
To make the $\sigma_h$ parameterization relevant across input sizes and dimensionalities,
we rescale the input domain $\X$ to be a unit hyper-cube and set $\sigma_h$ proportionally to the length of the diagonal of the domain $\sqrt{Q}$ (where $Q$ is the number of dimensions).

\textbf{Posterior inference and acquisition calculation.}
In BO we assume that $N$ pairs of inputs and outputs $\mat{X} = (\mat{x}_1, \ldots, \mat{x}_N)^\top$ and $\mat{F} = (\mat{f}_1, \ldots, \mat{f}_N)^\top$ have been collected.
To suggest the location for the next evaluation, we first need to infer the posterior distribution of the latent variables, which are $\mat{H}$ and $\mat{\theta}$ in LGP, and then search for the maximum of the acquisition function $\alpha(\mat{x})$.

Given the observed data, the probabilistic model of LGP is formulated as
\begin{align}
    \begin{split}
        \Prob{\mat{F} \given \mat{X}, \mat{H}, \mat{\theta}} = \mathcal{N}(\mat{F} | \mat{0}, \mat{K}),\\
        \Prob{\mat{H}} = \prod_{n=1}^N \mathcal{N}(\mat{h}_n | \mat{0}, \sigma_h^2\mathbb{I}),
    \end{split}
\end{align}
where $\mat{K}$ is the covariance matrix computed using a chosen kernel function $k(\cdot, \cdot)$ over the set of data points $\{\bar{\mat{x}}_n\}_{n=1}^N$, and $\bar{\mat{x}}_n$ is the concatenation of two vectors $(\mat{x}_n^\top, \mat{h}_n^\top)^\top$.
Because $\mat{h}_n^\top$ enters the kernel function non-linearly, it is clear that the posterior distribution $\Prob{\mat{H}, \mat{\theta} \given \mat{F}, \mat{X}}$ is intractable.
To ensure the quality of the acquisition function, usually, BO methods draw posterior samples of latent variables via Markov Chain Monte Carlo (MCMC) methods such as slice sampling~\cite{snoek2012practical} or Hamiltonian Monte Carlo~\citep{SimonEtAl1987}.
We follow this practice and provide details in the supplement.

With the approximate posterior samples $\{\mat{H}_i, \mat{\theta}_i\}_{i=1}^M$, we approximate the acquisition function with LGP in (\ref{eq:utility}) with Monte Carlo samples,
\begin{align}
    \alpha(\mat{x}_\ast) &\approx
    \frac{1}{M}\sum_{i=1}^M \hat{\alpha}(\mat{x}_*, \mat{H}_i, \mat{\theta}_i),
\end{align}
where $\hat{\alpha}(\mat{x}_*, \mat{H}_i, \mat{\theta}_i)$ is the acquisition function given the latent variables of LGP, which is closed-form for common acquisition functions such as expected improvement (EI) and upper confidence bound (UCB).

\begin{figure*}[ht]
    \centering
    \includegraphics[width=0.195\textwidth]{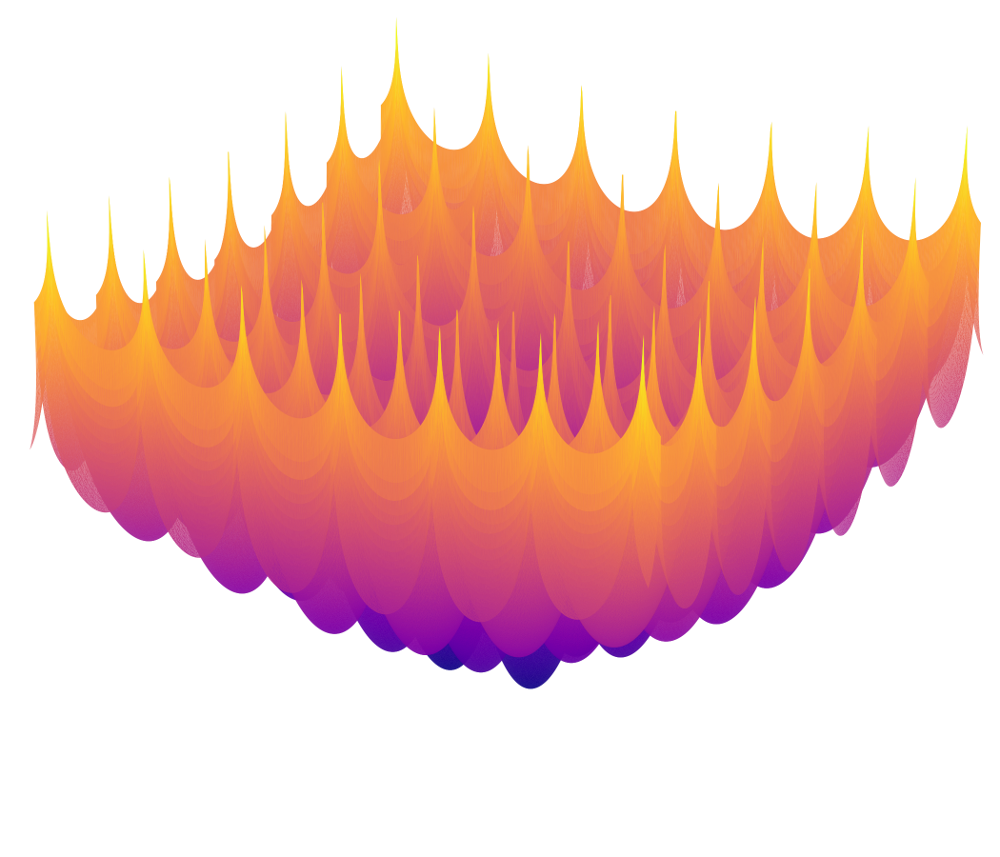}
    \includegraphics[width=0.195\textwidth]{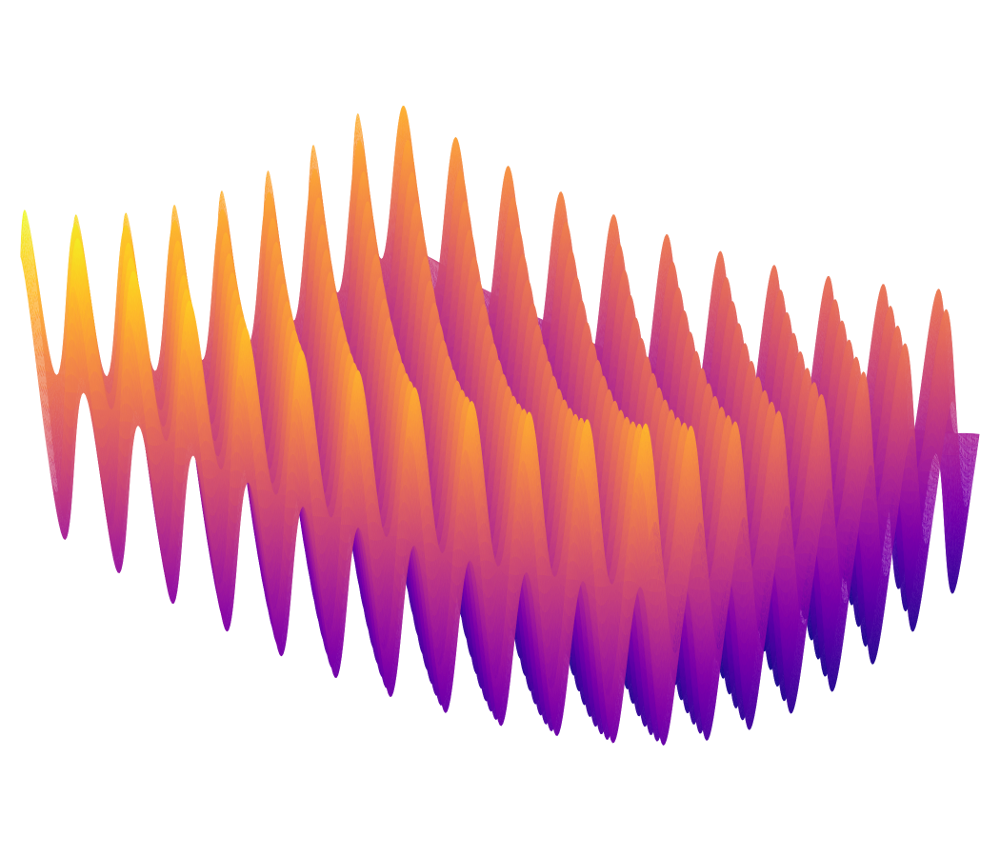}
    \includegraphics[width=0.195\textwidth]{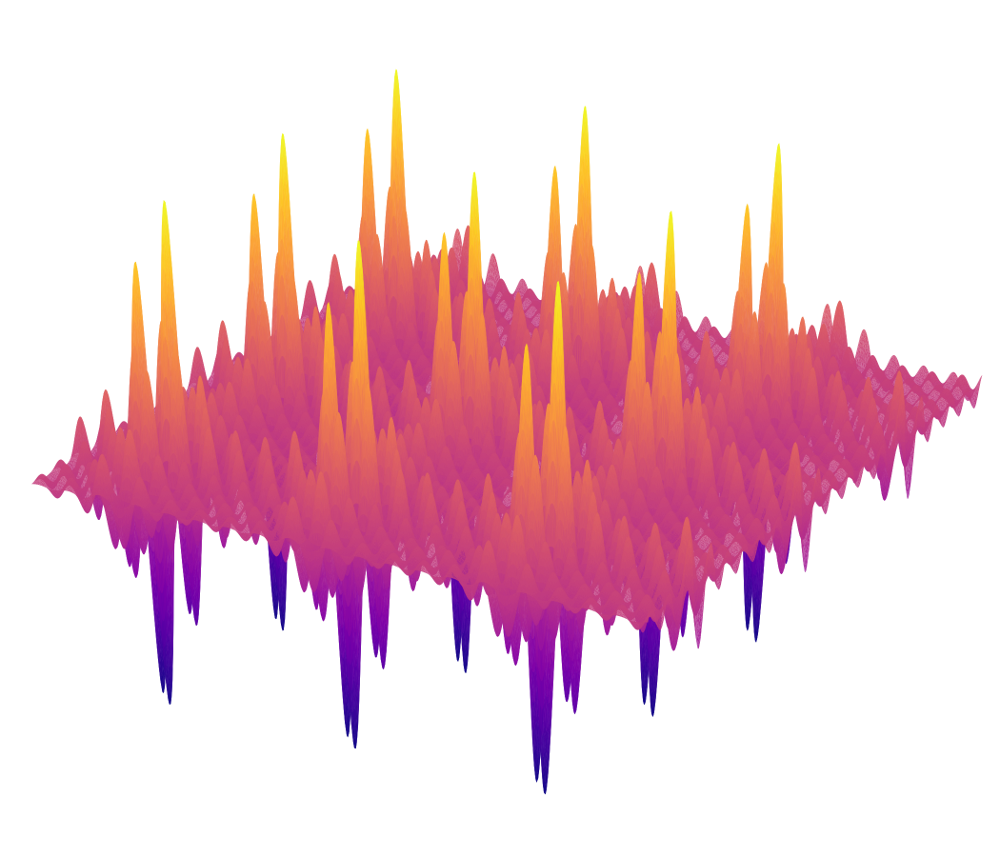}
    \frame{
    \includegraphics[width=0.195\textwidth]{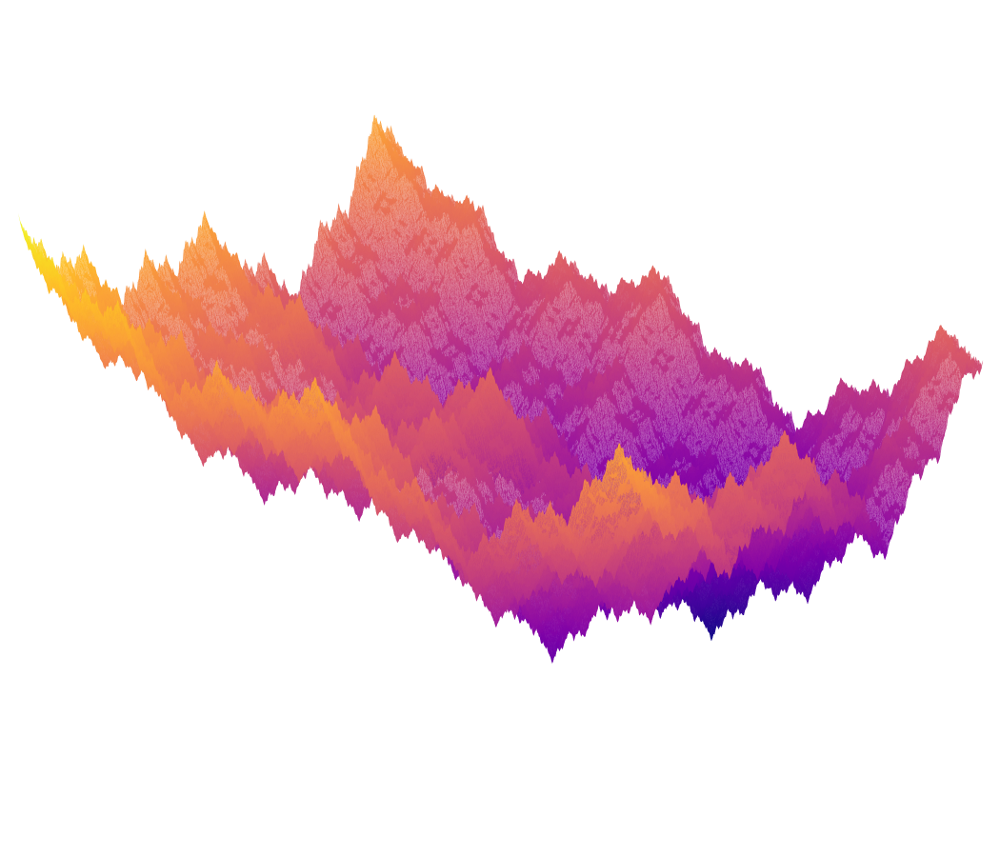}
    \includegraphics[width=0.195\textwidth]{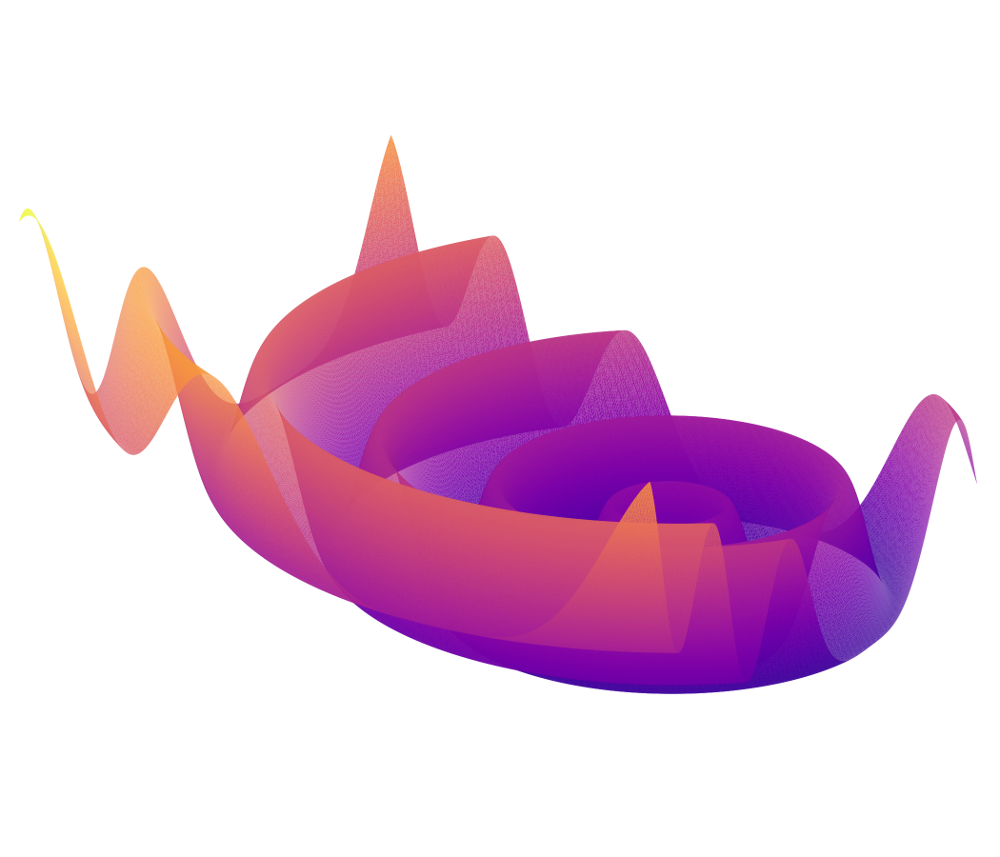}
    }
    \caption{
    Surface plots of the benchmark functions Cross In Tray, Griewank, Shubert, Weierstrass and Deflected Corrugated Spring~\cite{evalset}, from the left.
    The two right-most functions are available in multiple dimensionalities, where 8D and 10D is used in the experiments, respectively.
    \label{fig:functions_plots}
    }
\end{figure*}
\begin{figure}[ht]
    \centering
    \includegraphics[width=0.2\textwidth]{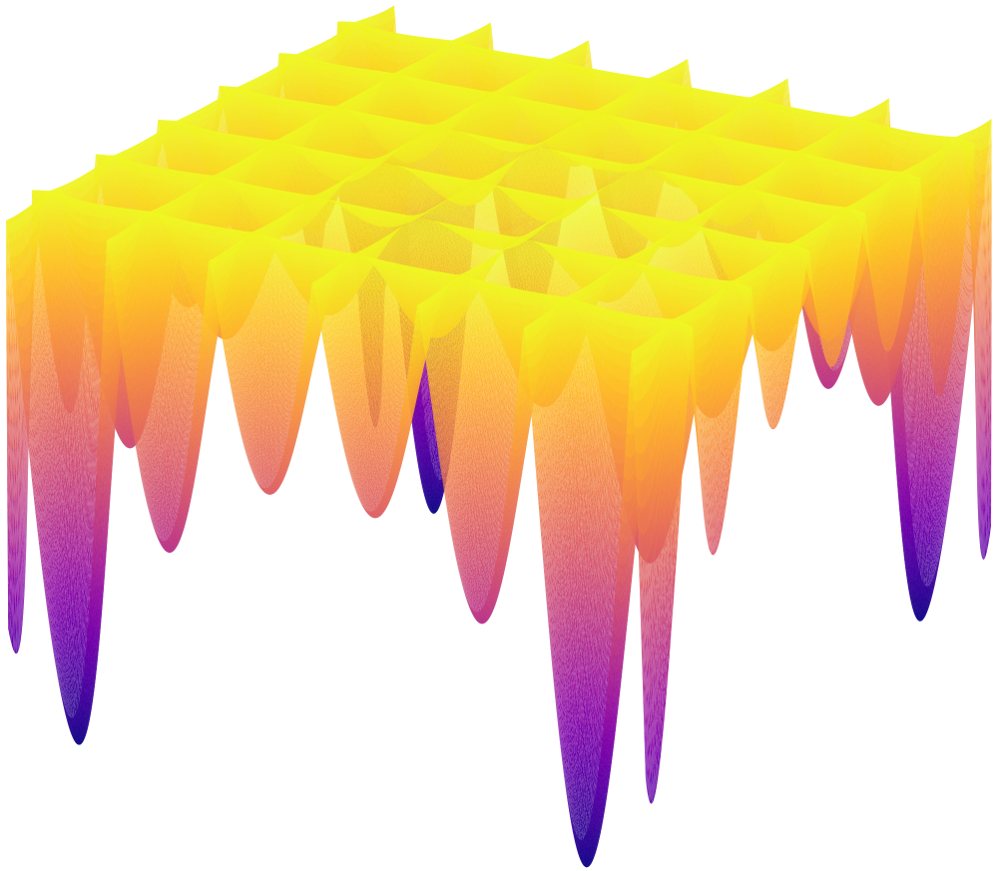}
    \hspace{0.5cm}
    \includegraphics[width=0.2\textwidth]{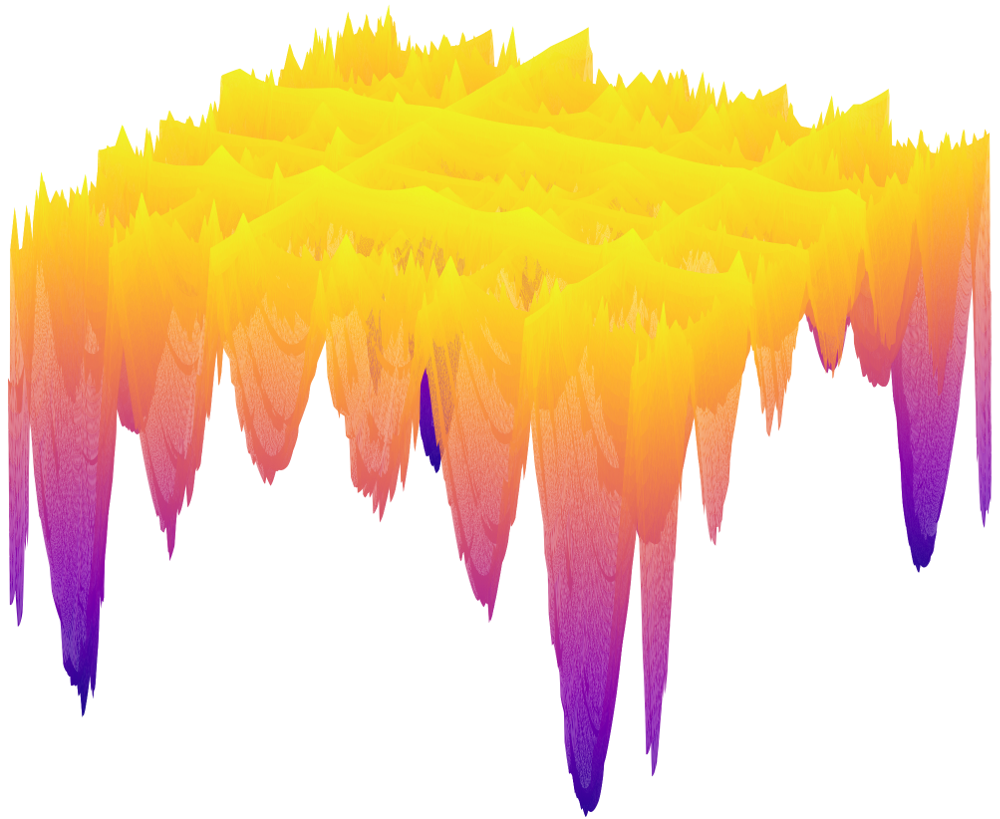}\\
    \vspace{-0.05cm}
    \includegraphics[width=0.235\textwidth]{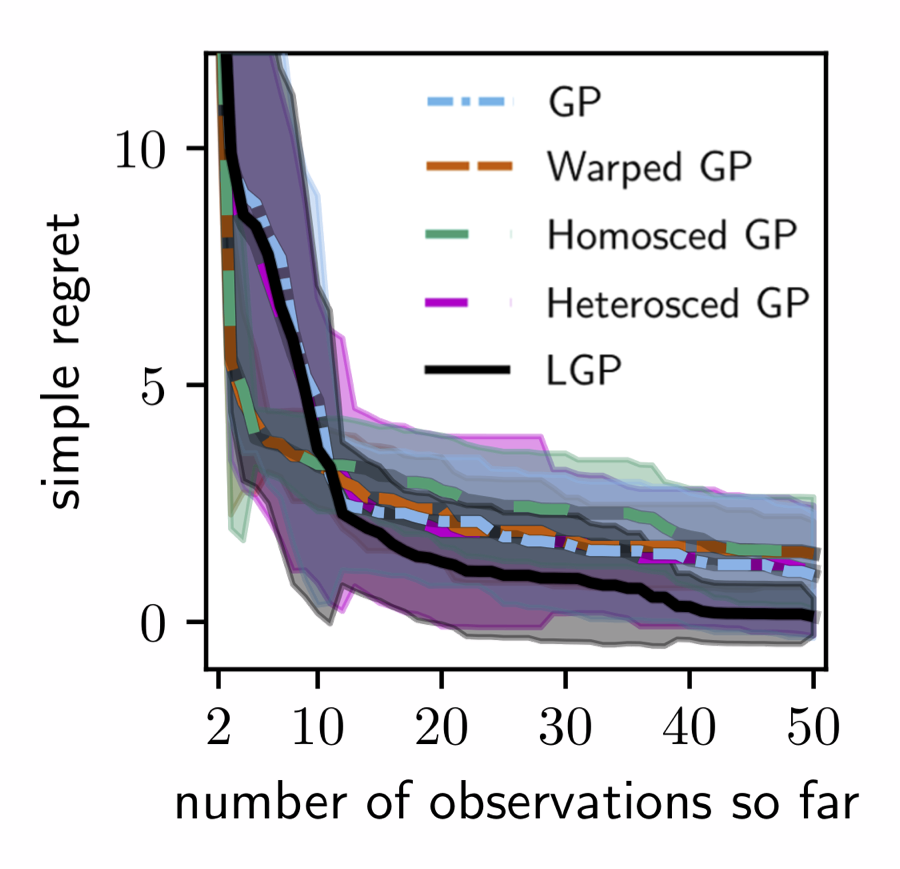}
    \includegraphics[width=0.235\textwidth]{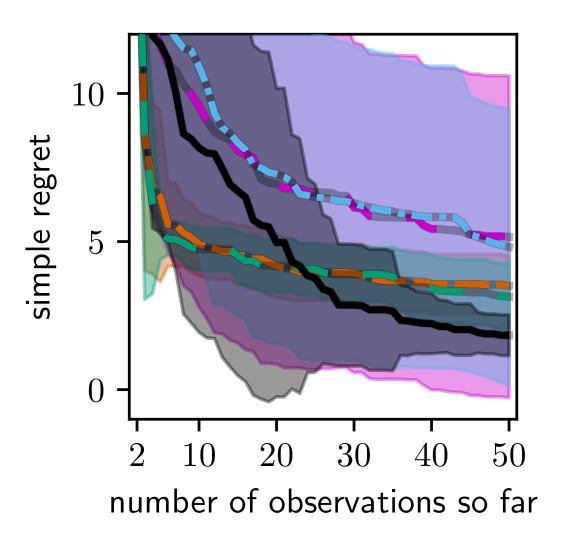}
    \caption{
    A comparison of experiments on the Holder Table benchmark~\cite{evalset} (left) and a corrupted version with added nonsmooth structure (right).
    We show plots of the respective functions and performance in terms of regret so far (over 20 repetitions).
    We show both mean regret (line) and standard deviation (shading).
    The nonsmooth structure is challenging for a noiseless GP to model and leads to a high variance in-between runs.
    Warped and homoscedasic GPs explain away the corruption, but their performance plateaus as no informative trends can be identified.
    LGP reliably identifies these trends and reliably finds good solutions.
    }

    \label{fig:corrupted_holder}
\end{figure}

\section{Related Work}
\label{section:related_work}
Performing BO on an objective function that is not well-behaved is very challenging.
Our method takes a Bayesian approach by incorporating a flexible noise distribution and
utilising Bayesian inference to assign the challenging details of the objective function to the noise distribution.
An alternative approach to this problem is to perform a model selection for the surrogate model, such that the choice of the surrogate model becomes a trade-off between the complexity of the model and the ability to locate the optimum under limited data, which has been explored in~\cite{malkomes2018automating}.
The approach uses a compositional kernel grammar from~\cite{duvenaud2013structure} to induce a space of GP models to choose from.
Although this and other model selection procedures~\cite{malkomes2016bayesian,duvenaud2013structure,grosse2012exploiting,gardner2017discovering} themselves have shown promise,
in addition to the computational overhead, the procedures are still reliant on the existence of suitable models in this space.
It remains challenging to handle cases where the objective function contains structure that is both hard to specify a-priori, and that is unhelpful in guiding the search to the optimum.

The idea of making use of noise models for dealing with model mismatch to noise-free data is not in itself new.
In~\cite{gramacy2012cases} it was shown that introducing noise in the modelling of noise-free computer experiments can lead
to models with better statistical properties such as predictive accuracy and coverage.
In that work, homoscedatic noise was addressed and used in a regression context.

In this paper we consider noise-free functions and address model misspecification of the function surrogate,
but many works have been done to make BO resilient to noisy experiments.
For example, robust noise distributions such as Student's t-distribution have been used to make BO more resilient to noise outliers~\cite{martinez2017robust,martinez2017practical}.
Approaches to address noisy experiments,
via the addition of likelihood functions,
can be combined with our approach.

Hierarchical surrogate models with input warpings have been proposed to tackle BO for non-stationary objective functions~\cite{snoek2014input,Oh2018BOCKB,calandra2016manifold}.
A particularly successful application is hyperparameter optimization for machine learning methods, in which the parameters are often presented in logarithmic scales.
In this case, the Beta cumulative distribution function, which only has two parameters, serves as a good warping function~\cite{snoek2014input}.
Such augmentation in surrogate models requires strong domain knowledge of the objective function, and one often still has to control the increased complexity of the surrogate model, which is orthogonal to our approach.

\section{Experiments}

In this section we will demonstrate the benefit of our approach empirically.
As the approach is motivated by robustness to the presence of challenging structures in the objective function,
we will test its ability to improve search efficiency on a range of functions exhibiting such structure.
Visual examples of functions with typical properties are shown in Figure~\ref{fig:functions_plots}.
As we will show,
our approach increases reliability in the search when faced with detrimental structure (see Figure~\ref{fig:corrupted_holder})
that has a large negative impact on traditional surrogates.

\paragraph{Baselines and metric}
We compare with and without function modulation (Section~\ref{sec:modulated_objectives})
-~implemented as in Section~\ref{sec:latent_gp_surrogate}~-
on a popular GP model setup for BO.
In addition, we compare the LGP against other methods of handling challenging structure in the objective function,
namely (i) a noiseless GP,
(ii) a GP with homoscedastic noise,
(iii) a GP with heteroscedastic noise and
(iv) a non-stationary,
Warped GP~\cite{snoek2014input}.

We follow the standard practice to compare across benchmarks and provide the \emph{mean gap} estimated over 20 runs as in~\cite{malkomes2018automating}.
The gap measure is defined as $\frac{f(x_{\text{first}}) - f(x_{\text{best}})}{f(x_{\text{first}}) - f(x_{\text{optimum}})}$,
where $f(x_{\text{first}})$ is the minimum function value among the first initial random points,
$f(x_{\text{best}})$ is the best function value found within the evaluation budget and $f(x_{\text{optimum}})$ is the function's true optimum.
Methods are judged to have very similar or equivalent performance to the best performing if not significantly different,
determined by a two-sided paired Wilcoxon signed-rank test at 5\% significance~\cite{malkomes2018automating}.
We also report regret (with mean and standard deviation) in the supplement.

We use the Mat\'{e}rn 5/2 kernel for all surrogates, the expected improvement acquisition function (where not otherwise stated) and Bayesian hyperparameter marginalisation as in~\cite{snoek2012practical}.
For the maximization of the expected utility with respect to input location,
we use $\delta$-cover sampling, as in~\cite{de2012exponential}.
The Warped GP implementation and inference is from the Spearmint package~\cite{snoek2014input}.
For further details, we refer to the supplement.

\begin{table*}[t]
    \centering
    \caption{
    \label{table:z_table}
    Mean gap performance for various test functions; higher is better.
    The upper table shows the results after 50 objective function evaluations and the lower table after 100 evaluations.
    Due to computational cost, Warped GP results are only reported for 50 evaluations.
    Methods not significantly different from the best performing method with respect by a two-sided paired Wilcoxon signed-rank test at a 5\% significance level over 20 repetitions are shown in bold~\cite{malkomes2018automating}.
    For results in terms of regret, see the supplement.
    }
    \resizebox{\textwidth}{!}{%
    \begin{tabular}{lrrlrr|rrr}
        \toprule
        {Benchmark} & Evals & Dim & \multicolumn{1}{c}{Properties} & \multicolumn{1}{c}{GP} & \multicolumn{1}{c}{Warped GP} & \multicolumn{1}{c}{Homosced GP} &  \multicolumn{1}{c}{Heterosced GP} & \multicolumn{1}{c}{LGP} \\
        \midrule
        Hartmann           & 50 & 6 & boring     &             \textbf{0.959} &     0.537 &   0.881 &     \textbf{0.973} &     \textbf{0.937} \\
        Griewank           & 50 & 2  & oscillatory             &           \textbf{0.914} &     0.493 &   0.752 &   \textbf{0.913} &       \textbf{0.897} \\
        Shubert            & 50 & 2 & oscillatory &            0.378 &     0.158 &   \textbf{0.378} &     \textbf{0.480} &     \textbf{0.593} \\
        Ackley $[-10, 30]^d$ & 50 & 2  & complicated, oscillatory                    &        \textbf{0.924} &     0.274 &   \textbf{0.892} &       \textbf{0.912} &   \textbf{0.927} \\
        Cross In Tray      & 50 & 2  & complicated, oscillatory             &            \textbf{0.954} &     0.385 &   0.929 &     \textbf{0.977}  &     \textbf{0.945} \\
        Holder table       & 50 & 2 & complicated, oscillatory         &            0.939 &     0.896 &   0.900 &     0.931 &          \textbf{0.993} \\
        Corrupted Holder Table & 50 & 2 & complicated, oscillatory &        0.741 &     0.798 &   0.826 &      0.729 &      \textbf{0.896} \\
        \midrule
        Branin01           & 100     & 2 &  none            &               \textbf{1.000} &&          \textbf{1.000} &                  \textbf{1.000} &            \textbf{1.000} \\
        Branin02           & 100     & 2 &  none            &               \textbf{0.991} &&   0.964 &          \textbf{0.990} &   \textbf{0.981} \\
        Beale              & 100     & 2 &  boring        &              \textbf{0.987} &&   \textbf{0.982} &         \textbf{0.987} &   \textbf{0.988} \\
        Hartmann           & 100     & 6 &  boring            &               \textbf{0.987} &&   0.947 &           \textbf{0.984} &   \textbf{0.979} \\
        Griewank           & 100     & 2 &  oscillatory            &              \textbf{0.967} &&   0.875 &         \textbf{0.969} &     \textbf{0.946} \\
        Levy               & 100     & 2 &  oscillatory            &               \textbf{0.997} &&   \textbf{0.999} &        \textbf{0.998} &      \textbf{0.998} \\
        Deflected Corrugated Spring & 100 & 10 &  oscillatory    &               0.347 &&   \textbf{0.840} &      0.406 &        0.697 \\
        Shubert $[-10, 10]^d$           & 100     & 2 &  oscillatory            &               0.510 &&   0.511 &          \textbf{0.672} &    \textbf{0.877} \\
        Weierstrass        & 100     & 8 &  complicated            &               0.600 &&   \textbf{0.704} &         0.577 &     0.625 \\
        Cross In Tray      & 100     & 2 &  complicated, oscillatory            &               \textbf{1.000} &&            0.995 &               \textbf{1.000} &               \textbf{1.000} \\
        Holder Table       & 100     & 2 &  complicated, oscillatory            &               0.971 &&            0.963 &                     \textbf{0.964} &         \textbf{1.000} \\
        Ackley $[-10, 30]^d$ & 100     & 2 &  complicated, oscillatory            &               \textbf{0.971} &&            0.914 &          \textbf{0.980} &                    \textbf{0.974} \\
        Ackley $[-10, 30]^d$ & 100     & 6 &  complicated, oscillatory            &               0.459 &&            \textbf{0.789} &                      0.442 &        \textbf{0.712} \\
        Corrupted Holder Table  & 100 & 2 & complicated, oscillatory            &              0.844 &&   0.889 &        0.822 &      \textbf{0.918} \\
        Corrupted Exponential & 100 & 8 &   complicated, oscillatory          &               0.580 &&   \textbf{0.847} &          0.581 &    \textbf{0.806} \\
        \midrule
        HPO: NN Boston          & 100     & 9 &  unknown  &           \textbf{0.720} &&   \textbf{0.761} &      \textbf{0.810} &        \textbf{0.770} \\
        HPO: NN Climate Model Crashes & 100 & 9 &  unknown  &          0.629 &&   \textbf{0.717} &        \textbf{0.683} &      \textbf{0.678} \\
        Active learning: Robot Pushing & 100 & 4 &  unknown  &          0.877 &&   0.745 &        \textbf{0.907} &      \textbf{0.932} \\
        \bottomrule
    \end{tabular}
    }
    \medskip
\end{table*}

\begin{figure*}[ht]
    \centering
    \includegraphics[width=0.31\textwidth]{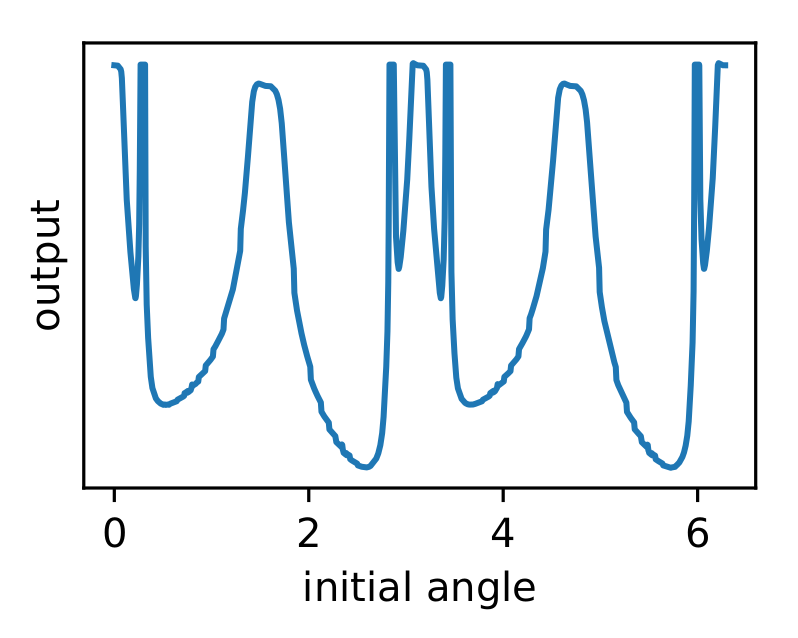}
    \includegraphics[width=0.31\textwidth]{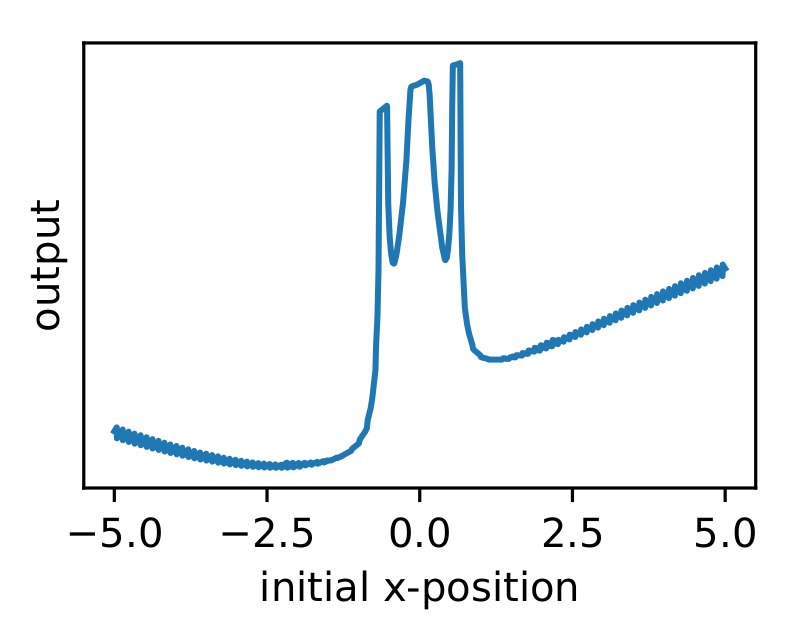}
    \includegraphics[width=0.335\textwidth]{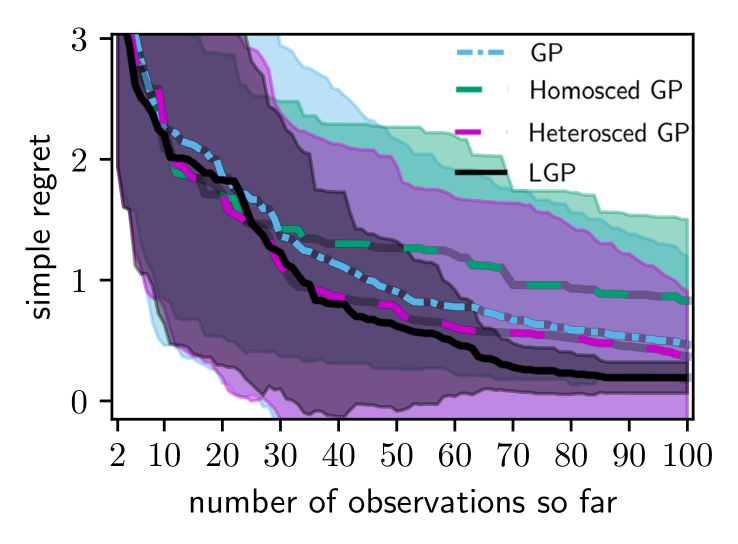}
    \caption{
    Active learning: Robot pushing.
    The objective function to be optimized takes as input the pushing action of a robot within a simulation,
    and outputs the distance of the pushed object to the goal location.
    The two plots on the left show that the task surface, resulting from the dynamic system, is nonsmooth and non-trivial.
    The right plot shows the mean regret and standard deviations for the different surrogates.
    As can be seen, the LGP found good solutions with low variance, improving reliability of the search.
    The 1D slices of the 4D function (the two from the left) was generated by fixing the initial y-position (param.) to the one of the goal position,
    the simulation steps (param.) to the center of its domain, and varying the initial angle (param.) or the x-position (param.), respectively,
    while keeping the other fixed at zero.
    Slicing the 4D function differently produced similar nonsmooth response curves.\\
    \label{fig:real_world}
    }
\end{figure*}

\paragraph{Benchmark datasets}
We perform the comparisons on benchmarks from~\cite{evalset,tim_head_2018_1207017} using the default domains provided by respective benchmark, detailed in the supplement.
In addition, problems are marked with the descriptive properties given in~\cite{evalset} and in the supplement that can reflect the relative difficulty of the task.

\paragraph{Priors on the latent input variables}
The prior $\Prob{\mat{h}_n} = \Gaussian{\mat{0}, \sigma_h^2\mathbb{I}}$ can be parameterized in relation to the relative scale of the characteristics to be ignored.
We specify the function prior over the product space $\X \times \Z$ using a kernel with common parameters for $\mat{x}_n$ and $\mat{h}_n$.
Thus, the standard deviation of the prior $\sigma_h$ relates directly to distances in the $\X$-direction.
When domain-specific knowledge is available, $\Prob{\mat{H}}$ may be specified at an appropriate scale.
However, we often do not have access to such knowledge.
In all our experiments, we adopt a hierarchical prior approach whereby $\sigma_h$ is sampled uniformly from a small candidate set at each evaluation.
Specifically, $\sigma_h \sim \Uniform{\Set{0.1 d, 0.01 d, 0}}$ where $d = \sqrt{Q}$, the length of the diagonal of the unit Q-dimensional hypercube.
We found that this approach performed well empirically and is applied consistently across all our experiments where not otherwise specified.
A choice of $\sigma_h \to 0$ corresponds to a noiseless GP without latent covariates.

\paragraph{Evaluation on benchmark suite}
Table~\ref{table:z_table} presents results across a wide range of benchmark functions consisting of the SigOpt benchmark suite~\cite{evalset}.
Three additional real-world benchmarks~\cite{tim_head_2018_1207017,malkomes2018automating,kaelbling2017learning} are included in the bottom section of the table.
The benchmarks from~\cite{evalset} are popular functions used in both black-box optimization as well as classic optimization literature.
As of the focus of the paper, benchmarks from the literature exhibiting challenging properties such as oscillatory local structures were included,
in addition to simpler functions for reference.

In general, the noise-free, homoscedastic and heteroscedastic noise GPs tend to either share best place with the LGP or be outperformed by
it.
The Warped GP, which warps the input space to obtain a tight fit to the data, consistently struggle with the complicated and oscillatory benchmarks.
On some benchmarks there are large differences in favour of using noisy surrogates on the noise-free benchmarks.
Such an example is Ackley 6D, which in the dataset is described as \enquote{technically oscillatory, but with such a short wavelength that its behavior probably seems more like noise}~\cite{evalset}.
Another example is the Shubert function, which has multiple sharp local optima surrounded by large oscillations.
On 2 of 18 benchmarks, the LGP was not best (or within the two-sided Wilcoxon test), but instead the homoscedastic GP.
These functions were Weierstrass, which has a homoscedastic characteristic (see Figure~\ref{fig:functions_plots}),
and Deflected Corrugated Spring, on both of which the LGP obtained the second highest mean gap.
In contrast to the LGP and the heteroscedastic GP, the noise model of the homoscedastic GP sometimes hurt performance in relation to the noise-free GP.
Given the black-box nature of functions in BO, it is important that the surrogate noise model 'turns off' adequately when not needed.
The heteroscedastic GP provided significant benefit on two benchmarks over the GP,
whereas the LGP provided such benefit on \emph{eight} benchmarks.

\paragraph{Real-world} Apart from widely used synthetic functions,
we also compare our method on three real-world problems.
The results are shown in the last three rows of Table~\ref{table:z_table}.
One of the benchmarks is an active learning task of a robot pushing a box within a simulation.
As we show in Figure~\ref{fig:real_world}, the benchmark's response surface is both nonsmooth and oscillatory.
The LGP reliably found good solutions on the benchmark, while the other surrogates sometimes failed, resulting in high variances.
The homoscedastic GP performed the worst,
which we suspect is due to the nonsmooth and heavily oscillatory structures forces a high global noise level,
which may lead to failure in utilising informative structure in other regions.

\paragraph{Other aquisition functions} We suggest that the problem with structures challenging to model is relevant to address irrespective of the acquisition function.
To confirm that the method is applicable also using other acquisition functions we ran the Corrupted Exponential benchmark
using both Expected Improvement (EI) and Lower Confidence Bound (LCB) with the default exploration weight ($=2.0$) from GPyOpt~\cite{gpyopt2016}.
As can be seen in Table~\ref{table:z_table}, in the case of EI, the GP and the heteroscedastic GP performed worse than the homoscedastic GP and the LGP.
The homoscedastic GP achieved the highest mean gap, but the difference was not significant under the Wilcoxon test to the LGP which obtained a similar mean gap.
Using the LCB acquisition function the performance for the homescedastic GP decreased to $0.818$ and the LGP increased to $0.858$,
and their difference in rank in favour of the LGP was significant under the test.
The heteroscedastic GP increased using LCB to $0.797$ and the GP to $0.751$, remaining as worst performing.

As the experimental evaluation demonstrates,
our suggested approach for handling challenging structures in the objective function consistently improved reliability and performance
over the traditional surrogate on a wide range of benchmarks.
Importantly, on benchmarks where the extended methodology were not needed the performance aligned with that of the traditional surrogate.
When it was needed, it was shown to often have a large positive impact on overall efficiency of the search.

\section{Conclusion}
We have presented an approach to Bayesian Optimization where the surrogate model is alleviated from needing to explain the observed objective function values perfectly,
which is challenging for complicated or nonsmooth functions.
Instead, we model the essential structure of the objective function that is well-behaved and leave the rest of the function details to be absorbed in a noise distribution.
We show experimentally how our approach is able to solve synthetic and real-world benchmarks with challenging local structures reliably.
Importantly our methodology can be applied to any surrogate model used for BO,
and the specific case addressed in the paper can be included in any Gaussian process-based surrogate.

\section{Acknowledgements}
This project was supported by Engineering and Physical Sciences Research Council (EPSRC) Doctoral Training Partnership,
the Hans Werthén Fund at The Royal Swedish Academy of Engineering Sciences,
EPSRC CDE (EP/L016540/1),
EPSRC CAMERA project (EP/M023281/1),
the German Federal Ministry of Education and Research (project 01 IS 18049 A),
and the Royal Society.

\balance
\bibliography{main}
\bibliographystyle{icml2020}

\newpage
\onecolumn
\appendix

\section{Appendix}

\subsection{$\Prob{\mat{h}_*}$ augmentation}

\begin{figure}[ht]
    \centering
    \captionsetup{width=.8\linewidth}
    \includegraphics[width=0.40\textwidth]{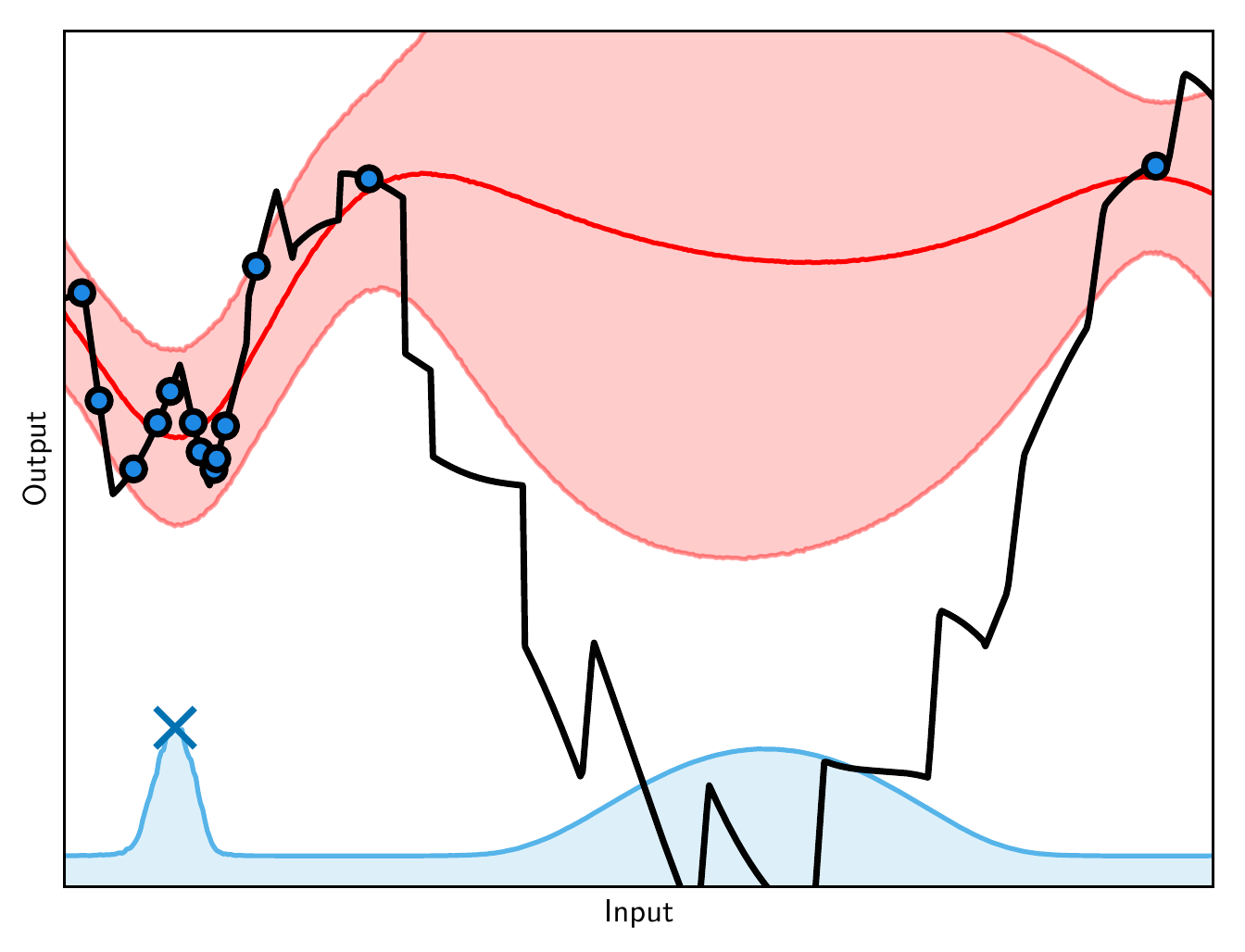}
    \includegraphics[width=0.40\textwidth]{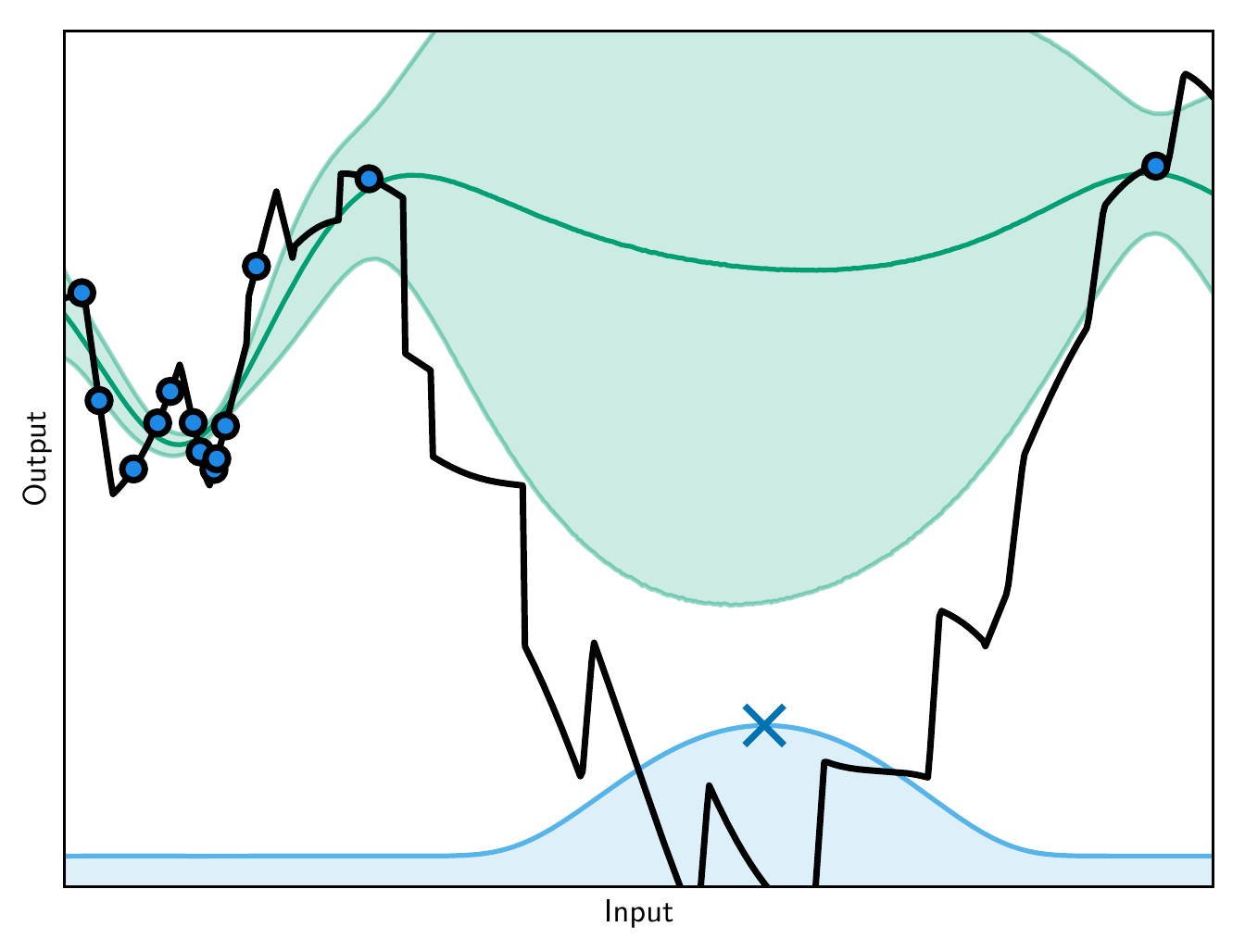}
    \caption{
    Ignoring irreducible uncertainty from $p(\mat{h}_\ast)$ in the acquisition.
    The effect of marginalizing $p(\mat{h}_\ast)$ as per standard versus collapsing the irreducible uncertainty of $h(\mat{x})$ as $p(\mat{h}_\ast) = \delta(\mat{h}_*)$
    is shown on the left and the right, respectively.
    Note that $p(\mat{H})$, associated with the observations, is marginalized in both cases.
    The acquisition (EI) and the next sample location is shown in blue.
    The posteriors are shown with mean and two standard deviations for display purposes, i.e. with estimated moments and approximated as Gaussian.
    Note that by incorporating variance induced from $p(\mat{h}_\ast)$, which cannot be reduced by acquiring data,
    the search can end up in a failure mode and get stuck by repeatedly evaluating in a region explained by high irreducible uncertainty in the objective function.
    By only considering model uncertainty in the function which is reducible by active sampling, i.e.~collapsing the predictive $p(\mat{h}_\ast)$ associated with new evaluations, the exploration of the function continues.
    }
    \label{fig:marg_z}
\end{figure}

The predictive distribution of the latent variable for new evaluations $\Prob{\mat{h}_*}$,
even at observed inputs locations,
will always by the i.i.d. assumption be equal to the prior.
As such, this source of uncertainty cannot be reduced by active sampling, nor does it reflect observational stochasticity which is made clear by the noiseless experiments.
Including it would simply include unhelpful artifacts in the decision upon where to collect data and in the worst case the search would be stuck, see~Figure~\ref{fig:marg_z}.
The importance of special treatment of similar uncertainty within active sampling has been noted in~\cite{gonzalez2017preferential}.

Noting the risk of ill-effects of simply marginalizing $\Prob{\mat{h}_*}$, one might be tempted to introduce statistical dependencies in the model
such that the belief about $\mat{h}_*$ associated with a new evaluation is updated from e.g. neighbouring observations.
However, such dependencies does not come for free, as they would by necessity limit the model's capability to explain away via $\mat{H}$.
For example, by constraining nuisance parameters to be similar within neighbourhoods of $\mathcal{X}$ the ability to be robust to discontinuities
such as step functions would be reduced.

We suggest to remove the effect $\mat{h}_*$ has on the predictive distribution of $\mat{f}_*$.
In the case of additive models this is easily achieved.
For example, in the GP regression context with noisy observations the predictive distribution of the noise-free latent function
is easily derived by removal of the noise variable after the noise has been considered in the data.
The reason why it is easy in the additive case is because the expectation is trivially separable, as $\mathbb{E}[g + h] = \mathbb{E}[g] + \mathbb{E}[h]$.
In this paper we consider the introduction of nuisance parameters in surrogates generally, including those with \emph{non-linear} interactions.
In the general non-linear case how to remove the interaction is an open question.
For these cases, when there is no available model-specific treatment, we propose the heuristic of collapsing the prior distribution of $\mat{h}_*$ into a Dirac delta distribution centered at its mode $\mat{h}_* = \bm{0}$,
which is consistent with the additive case.

\subsection{Further details on the objective function definition}
\begin{equation}
    f(\mat{x}) = g(\mat{x}, \mat{h}),\quad \mat{h} \sim \mathcal{N}(\mat{0}, \mathbb{I}) \label{eqn:objective},
\end{equation}
where $g$ is a well-behaved function that can be nicely modeled by a surrogate model of choice,
which is a Gaussian process (GP) in this paper,
and the vector-valued function $h(\mat{x})$ encodes the structures which the surrogate model struggles to capture.
Let's assume an uninformative prior distribution of $\mat{x}$ over $\X$, e.g.\ a uniform distribution $\mat{x} \sim \mathcal{U}(\X)$.
We denote the mean and variance of $h(\mat{x})$ under the prior distribution as $\mu_{\mat{h}} = \mathbb{E}_{\Prob{\mat{x}}}[h(\mat{x})]$ and $\Sigma_{\mat{h}} = \mathbb{E}_{\Prob{\mat{x}}}[(h(\mat{x})-\mu_{\mat{h}})(h(\mat{x})-\mu_{\mat{h}})^\top]$, respectively.
An important step to convert $\mat{h}$ into being part of a noise distribution is to treat it as random and independent of $\mat{x}$, i.e.\ $\mat{h}$ becomes an independent random variable with respect to each data point, just like a standard noise term.
In this paper, we use a normal distribution for $\mat{h}$, $\mat{h} \sim \mathcal{N}(\mu_{\mat{h}}, \Sigma_{\mat{h}})$.
In this way, the objective function becomes a function of two variables $g(\mat{x}, \mat{h})$, in which $\mat{h}$ is a random variable independent of $\mat{x}$ and which explain the data variance that cannot be explained by $\mat{x}$.
Note that the random variable $\mat{h}$ can be equivalently rewritten as $\mat{h} = \mu_{\mat{h}} + \mat{L}\bar{\mat{h}},\,  \Sigma_{\mat{h}}=\mat{L}\mat{L}^\top,\, \bar{\mat{h}} \sim \mathcal{N}(\mat{0}, \mathbb{I})$ and then the objective function becomes $g(\mat{x}, \mu_{\mat{h}} + \mat{L}\bar{\mat{h}})$.
As $g$ is a non-linear function inferred during BO, the linear transform can be absorbed into the formulation.

\subsection{Step function example}
\begin{minipage}{\linewidth}
    \makebox[\linewidth]{
    \centering
    \includegraphics[width=0.40\textwidth]{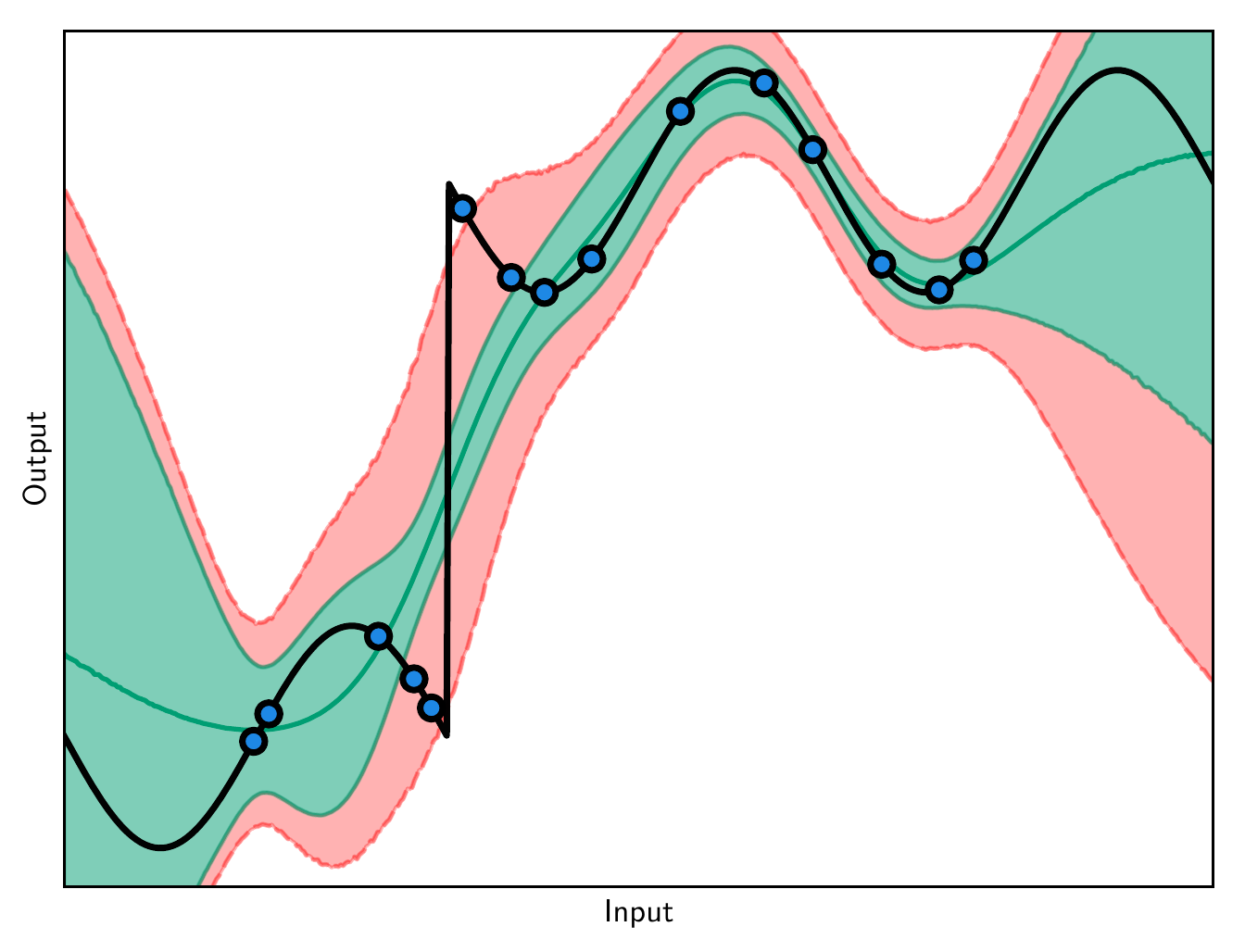}
    }
    \captionof{figure}{
    Robustness to nonsmooth structures.
    The modulated function posterior (of the LGP) is shown in green with mean and two standard deviations.
    The posterior of $f$ (with $p(\mat{h}_\ast)$ marginalized) is shown with two standard deviations from its mean in red.
    The true function is shown in black.
    Both posteriors are for display purposes approximated as Gaussian as in Figure~\ref{fig:marg_z}.
    Note how some structure of $f$ is ignored and treated as non-additive, heteroscedastic noise.
    }
    \label{fig:robustness}
\end{minipage}

\subsection{Model setup, inference and auxiliary optimization}
\label{appendix:approx_inference}

We use the Warped GP~\cite{snoek2014input} surrogate model with default settings as provided by the Spearmint library~\cite{spearmint}.
We use the Mat\'{e}rn 5/2 kernel for all surrogates and the model-marginalised expected improvement acquisition function as in~\cite{snoek_practical_2012}.
For all baselines we make us of hyper-parameter marginalisation via Markov Chain Monte Carlo (MCMC)~\cite{shahriari_taking_2016}.

For the noise-free, homoscedastic GP and Warped GP (all with few parameters) we use slice sampling, as recommended for BO in~\cite{snoek2012practical}
due to its automatic adjustment of the step size to match the local shape of the density function.
The heteroscedastic GP has a latent noise variance per observation.
Similarly, LGP has the latent inputs $\mat{H}$ associated with the observations, making inference impractical for both of these models using slice sampling as of comparably large dimensionalites.
For these models, in all BO experiments, we use Hamiltonian Monte Carlo with step size adaptation.
A burn-in of $30,000$ steps and a thinning rate of $x50$ (select every 50th value) were used, and $100$ posterior samples collected.
Step-size adaptation were made during $80\%$ of the burn-in phase, with a target acceptance probabilty of $0.75$ which is the center of asymptotically optimal rate for HMC~\cite{betancourt2014optimizing}.

Observed outputs are normalized to have zero mean and unit variance (standard normalization) at each iteration.
For the GP surrogate with and without the noise models as well using the latent input extension, we use a LogNormal(0, 1) prior for the lengthscale and noise variance parameters and use unit signal variance.
We use one dimension for the latent variables $\mat{h}_n$ in the comparisons with the baselines.
For the maximization of the expected utility with respect to input location,
we use $\delta$-cover sampling, as in~\cite{de2012exponential}, for all models (where in our case the expected utility is the model-marginalised expected improvement).
The sampling scheme works by iteratively sampling the utility more densely in $\mathcal{X}$ around the location of current highest obtained utility.
Specifically, we double the concentration of an uniform sample density at each auxiliary iteration by multiplication of each side of the current sampling hypercube by a factor of $2^{-\nicefrac{1}{Q}}$ for $30$ iterations.
Without loss of generality, all function domains are re-scaled to unit hypercubes for a consistent parameterization of priors across functions.
In all experiments we start from $2$ uniformly drawn initial observations and stop after $50$ and $100$ observations, respectively, as indicated in the experiments.

\subsection{Benchmarks}

\subsubsection{Property labels}
\label{appendix:property_labels}
For benchmark property labels, see Table~\ref{table:props}.

\begin{table}[h!]
    \begin{center}
        \caption{
        Function properties as defined in SigOpt~\cite{evalset}.
        }
        \scalebox{0.72}{
        \label{table:props}
        \begin{tabular}{ll}
            \toprule
            {Property} & Meaning \\
            \midrule
            boring               & A mostly boring function that only has a small region of action  \\
            oscillatory               & A function with a general trend and an short range oscillatory component  \\
            complicated               & These are functions that may fit a behavior, but not in the most obvious or satisfying way  \\
            \bottomrule
        \end{tabular}
        }
    \end{center}
\end {table}

\subsubsection{Domains}
\label{appendix:domains}
\begin{equation}
    \text{pow10}(x) = 10^x
\end{equation}
\begin{equation}
    \text{pow2int}(x) = \text{int}(2^x)
\end{equation}
~ \\
For benchmark domains, see Table~\ref{table:domains}.

\begin{table}[h]
    \begin{center}
        \caption{
        Benchmark function domains used as specified in SigOpt~\cite{evalset} 'evalset'.
        The domains of Corrupted Holder Table and Corrupted Exponential correspond to the ones of Holder Table and Exponential, respectively.
        }
        \vspace{0.3cm}
        \label{table:domains}
        \scalebox{0.72}{
        \begin{tabular}{lcll}
            \toprule
            {Benchmark} & Dim & Properties & Domain \\
            \midrule
            Branin01               & 2 &  none            &   $[[-5, 10], [0, 15]]$ \\
            Branin02               & 2 &  none            &      $ [[-5, 15], [-5, 15]]$    \\
            Powell Triple Log  & 12 & none                  &   $[-4, 1]^{12}$ \\
            Beale                 & 2 &  boring        &        $[-4.5, 4.5]^2$      \\
            Hartmann & 6 & boring                            &  $[0, 1]^6$ \\
            Griewank & 2 & oscillatory                  &  $[-50, 20]^2$ \\
            Shubert01 & 2 & oscillatory                  &  $[-10, 10]^2$  \\
            Levy13                  & 2 &  oscillatory            &  $[-10, 10]^2$  \\
            Cosine Mixture & 10 & oscillatory                     &  $[-1, 1]^{10}$      \\
            Drop-Wave & 10 & oscillatory                          &  $[-2, 5.12]^{10}$      \\
            Deflected Corrugated Spring & 10 & oscillatory        &  $[0, 7.5]^{10}$      \\
            Weierstrass & 8 & complicated                         &  $[-0.5, 0.2]^8$  \\
            Cross In Tray         & 2 &  complicated, oscillatory            &  $[-10, 10]^2$  \\
            Holder Table & 2 & complicated, oscillatory                  &    $[-10, 10]^2$  \\
            Ackley $[-10, 30]^d$   & 2 &  complicated, oscillatory            & $[-10, 30]^2$ \\
            Ackley $[-10, 30]^d$ & 6 & complicated, oscillatory                              & $[-10, 30]^6$   \\
            Corrupted Holder Table & 2 &  complicated, oscillatory            & $[-10, 10]^2$ \\
            Corrupted Exponential & 8 &  complicated, oscillatory            & $[-0.7, 0.2]^8$ \\
            NN Boston& 9 & unknown                           &   Table~\ref{table:nn}   \\
            NN Climate Model Crashes & 9 & unknown            &  Table~\ref{table:nn}        \\
            Robot Pushing & 4 & unknown            &  Table~\ref{table:robot}        \\
            \bottomrule
        \end{tabular}
        }
    \end{center}
\end{table}

\begin{table}[h!]
    \begin{center}
        \caption{
        Neural Network hyperparameter domains as from~\cite{tim_head_2018_1207017}.
        Categorical options are set as specified below.
        }
        \vspace{0.3cm}
        \label{table:nn}
        \scalebox{0.72}{
        \begin{tabular}{ll}
            \toprule
            {Parameter} & Domain \\
            \midrule
            hidden\_layer\_sizes   & pow2int([1.0, 10.0]) \\
            alpha & pow10([-5.0, -1]) \\
            batch\_size & pow2int([5.0, 10.0]) \\
            max\_iter & pow2int([5.0, 8.0]) \\
            learning\_rate\_init & pow10([-5.0, -1]) \\
            power\_t & [0.01, 0.99] \\
            momentum & [0.1, 0.98] \\
            beta\_1 & [0.1, 0.98] \\
            beta\_2 & [0.1, 0.9999999] \\
            \hline
            learning\_rate & constant \\
            solver & adam \\
            activation & relu \\
            nesterovs\_momentum & False \\
            \bottomrule
        \end{tabular}
        }
    \end{center}
\end{table}

\begin{table}[h!]
    \begin{center}
        \caption{
        Active learning task of 'robot pushing' as from~\cite{kaelbling2017learning}.
        Code is available at \mbox{https://github.com/zi-w/Max-value-Entropy-Search}.
        All instances of np.random.normal(0, 0.01) was replaced by np.random.normal(0, 1e-6) in push\_world.py to make the function virtually noise-free.
        The goal location was set to a fixed location for reproducibility, as specified below.
        }
        \label{table:robot}
        \scalebox{0.72}{
        \begin{tabular}{ll}
            \toprule
            {Parameter} & Domain \\
            \midrule
            $\text{robot}_x$ & [-5, 5] \\
            $\text{robot}_y$ & [-5, 5] \\
            $\theta_{\text{initial}}$ & [0, 2$\pi$] \\
            simulation steps & int([10., 300.]) \\
            \hline
            $\text{goal}_x$ & 3.0 \\
            $\text{goal}_y$ & 2.0 \\
            \bottomrule
        \end{tabular}
        }
    \end{center}
\end{table}

\subsubsection{Regret version of results}
For regret version of results table, see Table~\ref{table:regret}.

\subsubsection{Corrupted Holder Table and Corrupted Exponential}
\label{appendix:corrupted}

The following corruption functions in Figure~\ref{fig:func} was used for the benchmarks Corrupted Holder Table and Corrupted Exponential.
\emph{small\_corruption\_func} and \emph{large\_corruption\_func} was applied to the SigOpt benchmarks~\cite{evalset} Holder Table 2D and Exponential 8D, respectively.
The input dimensions are re-scaled to be between 0 and 1 before the corruption is applied.
The new function minimum of each function (due to the corruption) was estimated via $1e6$ uniformly drawn samples in the domains.

\begin{figure*}
    \begin{center}
        \begin{lstlisting}[language=Python, basicstyle=\tiny\ttfamily]

            import numpy as np
            from scipy import signal

            def corruption(x, a0, a1, a2, a3):
                assert np.all(x >= 0)
                assert np.all(x <= 1)

                a = 0.0 + 1.0 * signal.square(4 * 2 * np.pi * x)
                b = 0.5 + 0.5 * signal.square(4 * 2 * np.pi * x)
                base = a * b

                p0 = 0.3 * np.pi
                p1 = 0
                p2 = np.pi
                p3 = 0.5 * np.pi

                s0 = a0 * signal.sawtooth(p0 + 15 * 2 * np.pi * x)
                s1 = a1 * signal.sawtooth(p1 + 10 * 2 * np.pi * x)
                s2 = a2 * signal.sawtooth(p2 + 30 * 2 * np.pi * x)
                s3 = a3 * signal.sawtooth(p3 + 40 * 2 * np.pi * x)
                return base * (s0 + s1 + s2 + s3)

            def corrupt(func, bounds, f_min, f_max, corruption_func):
                f_range = f_max - f_min
                lower_limits = np.array([b[0] for b in bounds])
                upper_limits = np.array([b[1] for b in bounds])
                ranges = upper_limits - lower_limits
                def normalise(v):
                    return (v - lower_limits) / ranges
                return lambda v: func(v) + \
                    f_range * np.max([corruption_func(v_dim_norm) for v_dim_norm in normalise(v)])

            small_corruption_func = lambda x: corruption(x, a0=-0.03, a1=0.05, a2=0.08, a3=0.03)
            large_corruption_func = lambda x: corruption(x, a0=-0.03, a1=0.20, a2=0.16, a3=0.06)
        \end{lstlisting}
        \caption{
        Corruption functions in Python.
        }
        \label{fig:func}
    \end{center}
\end{figure*}

\begin{table*}[h!]
    \centering
    \caption{
    \label{table:regret}
    Regret for various test functions; lower is better.
    The upper table shows the results after 50 objective function evaluations and the lower table after 100 evaluations.
    Due to computational cost, Warped GP results are only reported after 50 evaluations.
    }
    \resizebox{\textwidth}{!}{%
    \begin{tabular}{lrrrrlrrrrr}
        \toprule
        {Benchmark} & Evals & Dim & Func Max & Func Min & Initial regret & \multicolumn{1}{c}{GP} & \multicolumn{1}{c}{Warped GP} & \multicolumn{1}{c}{Homosced GP} & \multicolumn{1}{c}{Heterosced GP} & \multicolumn{1}{c}{LGP} \\
        \midrule
        Hartmann  & 50 & 6    &         0.00 &        -3.32 &     2.764~$\pm$~0.490 &     0.117~$\pm$~0.109 &     1.360~$\pm$~0.783 &     0.343~$\pm$~0.575 &   0.074~$\pm$~0.081 &    0.190~$\pm$~0.436 \\
        Griewank  & 50 & 2    &         3.19 &         0.00 &     1.151~$\pm$~0.441 &     0.104~$\pm$~0.230 &     0.379~$\pm$~0.179 &     0.220~$\pm$~0.150 &  0.105~$\pm$~0.101 &    0.102~$\pm$~0.081 \\
        Shubert  & 50 & 2   &       210.45 &      -186.73 &   183.2~$\pm$~9.820 &  114.6~$\pm$~51.93 &  129.3~$\pm$~31.52 &  113.1~$\pm$~47.77 & 95.70~$\pm$~60.38 &  75.34~$\pm$~66.56 \\
        Ackley $[-10, 30]^d$ & 50 & 2       &        22.27 &         0.00 &    16.45~$\pm$~4.202 &     1.391~$\pm$~2.101 &    12.79~$\pm$~3.550 &     1.641~$\pm$~0.723 &   1.282~$\pm$~2.481 &    1.093~$\pm$~0.741 \\
        Cross In Tray & 50 & 2 &        -0.26 &        -2.06 &     0.393~$\pm$~0.158 &     0.018~$\pm$~0.052 &     0.195~$\pm$~0.123 &     0.026~$\pm$~0.032 &   0.009~$\pm$~0.038 &    0.026~$\pm$~0.062 \\
        Holder Table & 50 & 2 &         0.00 &       -19.21 &    15.51~$\pm$~2.571 &  0.983~$\pm$~1.329 &  1.433~$\pm$~0.687 &  1.451~$\pm$~1.182 & 1.081~$\pm$~1.376 &    0.112~$\pm$~0.398 \\
        Corrupted Holder Table & 50 & 2 &         3.46 &       -20.99 &    18.37~$\pm$~2.444 &  4.816~$\pm$~4.671 &  3.508~$\pm$~0.983 & 3.138~$\pm$~1.123 & 5.157~$\pm$~5.433 &    1.836~$\pm$~0.690 \\
        \midrule
        Branin01           & 100     & 2    &       308.13 &         0.40 &        6.975~$\pm$~5.110 &    0.001~$\pm$~0.000 &&   0.001~$\pm$~0.001 &  0.001~$\pm$~0.001 &    0.000~$\pm$~0.000 \\
        Branin02                        & 100 & 2     &       506.98 &         5.56 &      27.666~$\pm$~28.450 &  0.253~$\pm$~0.500 &&    0.736~$\pm$~0.690 &  0.253~$\pm$~0.501 &  0.380~$\pm$~0.572 \\
        Beale           & 100     & 2     &    181853.61 &         0.00 &  1152.964~$\pm$~3639.596 &  0.288~$\pm$~0.302 &&    0.259~$\pm$~0.210 &  0.241~$\pm$~0.253 &  0.250~$\pm$~0.251 \\
        Hartmann                        & 100 & 6     &         0.00 &        -3.32 &        2.764~$\pm$~0.490 &    0.038~$\pm$~0.082 &&    0.162~$\pm$~0.471 & 0.044~$\pm$~0.066 &    0.058~$\pm$~0.100 \\
        Griewank                        & 100 & 2     &         3.19 &         0.00 &        1.151~$\pm$~0.441 &    0.028~$\pm$~0.020 &&    0.099~$\pm$~0.054 &  0.032~$\pm$~0.023 &    0.055~$\pm$~0.056 \\
        Levy                          & 100 & 2     &       454.13 &         0.00 &      58.26~$\pm$~37.30 &    0.289~$\pm$~1.018 &&    0.026~$\pm$~0.028 &  0.205~$\pm$~0.727 &    0.034~$\pm$~0.025 \\
        Shubert $[-10, 10]^d$                      & 100 & 2     &       210.45 &      -186.73 &      183.2~$\pm$~9.820 &  90.09~$\pm$~56.44 &&  88.72~$\pm$~50.59 & 60.34~$\pm$~58.95 &  22.70~$\pm$~45.82 \\
        Deflected Corrugated Spring    & 100 & 10     &        25.87 &        -1.00 &        5.919~$\pm$~1.364 &    3.921~$\pm$~1.848 &&    0.944~$\pm$~0.574 & 3.411~$\pm$~1.243 &    1.797~$\pm$~0.880 \\
        Weierstrass                     & 100 & 8     &       144.00 &       112.00 &       12.90~$\pm$~1.430 &    5.070~$\pm$~0.997 &&    3.790~$\pm$~0.767 &  5.401~$\pm$~1.320 &    4.798~$\pm$~1.436 \\
        Cross In Tray                   & 100 & 2     &        -0.26 &        -2.06 &        0.393~$\pm$~0.158 &    0.000~$\pm$~0.000 &&    0.001~$\pm$~0.002 &  0.000~$\pm$~0.000 &    0.000~$\pm$~0.000 \\
        Holder Table                    & 100 & 2     &         0.00 &       -19.21 &       15.51~$\pm$~2.571 &    0.459~$\pm$~1.048 &&    0.495~$\pm$~1.079 & 0.591~$\pm$~1.176 &    0.005~$\pm$~0.006 \\
        Ackley $[-10, 30]^d$                          & 100 & 2     &        22.27 &         0.00 &       16.45~$\pm$~4.202 &    0.480~$\pm$~0.720 &&    1.318~$\pm$~0.556 &  0.347~$\pm$~0.597 &    0.395~$\pm$~0.332 \\
        Ackley $[-10, 30]^d$                        & 100 & 6     &        22.27 &         0.00 &       19.89~$\pm$~0.809 &   10.92~$\pm$~6.656 &&    4.191~$\pm$~0.751 &  11.20~$\pm$~6.438 &    5.819~$\pm$~5.449 \\
        Corrupted Holder Table          & 100 & 2     &         3.46 &       -20.99 &       18.37~$\pm$~2.444 &    2.921~$\pm$~3.728 &&    2.034~$\pm$~0.781 & 3.396~$\pm$~3.798 &    1.474~$\pm$~0.342 \\
        Corrupted Exponential         & 100 & 8     &        -0.04 &        -0.99 &        0.391~$\pm$~0.101 &    0.172~$\pm$~0.109 &&    0.057~$\pm$~0.016 &  0.165~$\pm$~0.087 &    0.074~$\pm$~0.038 \\
        \midrule
        HPO: NN Boston          & 100 & 9     &         5.00 &        -0.85 &        0.886~$\pm$~1.199 &    0.023~$\pm$~0.012 &&    0.023~$\pm$~0.009 &  0.019~$\pm$~0.013 &    0.021~$\pm$~0.010 \\
        HPO: NN Climate Model Crashes & 100 & 9  &         5.00 &         0.11 &        0.161~$\pm$~0.099 &    0.047~$\pm$~0.017 &&    0.035~$\pm$~0.017 &  0.040~$\pm$~0.021 &    0.039~$\pm$~0.017 \\
        Active learning: Robot Pushing & 100 & 4  &         unknown &         0.00 &        3.783~$\pm$~1.842 &    0.468~$\pm$~0.729 &&    0.829~$\pm$~0.673 &    0.369~$\pm$~0.537 &    0.193~$\pm$~0.128 \\
        \bottomrule
    \end{tabular}
    }
    \medskip
\end{table*}
    
\end{document}